\documentclass{article}

\usepackage{iclr2027_conference,times}
\iclrfinalcopy
\usepackage{fontawesome5}

\usepackage[utf8]{inputenc}
\usepackage[T1]{fontenc}
\usepackage{hyperref}
\usepackage{url}
\usepackage{booktabs}
\usepackage{amsfonts}
\usepackage{amsmath,amssymb}
\usepackage{nicefrac}
\usepackage{microtype}
\usepackage{xcolor}
\usepackage{graphicx}
\usepackage{multirow}
\usepackage{wrapfig}
\usepackage{todonotes}
\usepackage{enumitem}
\AddToHook{env/table/begin}{\setlength{\belowcaptionskip}{6pt}}
\AddToHook{env/table*/begin}{\setlength{\belowcaptionskip}{6pt}}
\AddToHook{env/wraptable/begin}{\setlength{\belowcaptionskip}{6pt}}
\newcommand{\R}{\mathbb{R}}

\definecolor{revblue}{HTML}{1976D2}

\newcommand{\std}[1]{\text{\scriptsize{($\pm$#1)}}}

\definecolor{purple}{HTML}{C2185B}

\usepackage{tikz}
\usetikzlibrary{arrows.meta}
\usepackage{amsmath,bm}
\usepackage{pifont}
\usepackage{graphicx}
\usepackage{fix-cm}

\hypersetup{pdftitle={SkillEval: Learning Interpretable Ability Profiles of LLMs via Cognitive Diagnosis Models}, pdfauthor={Berke Arda, Peng Cui, Qiaoyuan Zheng, Rudolf Debelak, Mubashara Akhtar, Mrinmaya Sachan}, colorlinks=true, linkcolor=blue!60!black, citecolor=blue!60!black, urlcolor=blue!60!black}
\title{\fontsize{16}{20}\selectfont\textsc{SkillEval}: Learning Interpretable Ability\\Profiles of LLMs via Cognitive Diagnosis Models}

\author{\hspace*{-\tabcolsep}\mdseries\begin{tabular}[t]{@{}p{0.47\textwidth}p{0.47\textwidth}@{}} \textbf{Berke Arda}\thanks{Equal contribution.} & \textbf{Peng Cui}\footnotemark[1] \\ Department of Computer Science & Department of Computer Science \\ ETH Zurich & ETH Zurich \\ \texttt{bearda@ethz.ch} & \texttt{pencui@ethz.ch} \\[2.5ex] \textbf{Qiaoyuan Zheng} & \textbf{Rudolf Debelak} \\ Department of Computer Science & Machine Learning and Optimization Lab \\ ETH Zurich & EPFL \\ \texttt{zqiaoyuan@ethz.ch} & \texttt{rudolf.debelak@epfl.ch} \\[2.5ex] \textbf{Mubashara Akhtar} & \textbf{Mrinmaya Sachan} \\ ETH AI Center & Department of Computer Science \\ ETH Zurich & ETH Zurich \\ \texttt{mubashara.akhtar@ai.ethz.ch} & \texttt{msachan@ethz.ch} \end{tabular}}

\ifdefined\pdfsuppressptexinfo\pdfsuppressptexinfo=-1\fi

\begin{document}
\maketitle
\lhead{}\renewcommand{\headrulewidth}{0pt}
\vspace{-1.2em}\noindent{\small\begin{tabular}{@{}p{0.47\textwidth}p{0.47\textwidth}@{}}\faGithub~\href{https://github.com/berkearda/skilleval}{\texttt{github.com/berkearda/skilleval}} & \faGlobe~\href{https://berkearda.github.io/skilleval/}{\texttt{berkearda.github.io/skilleval}}\end{tabular}}\par\vspace{0.6em}
\begin{abstract}

Current evaluation of LLMs on benchmarks typically aggregates performance over individual examples into a single scalar score. 
This approach is overly coarse and ignores the diverse scope of skills covered by benchmark datasets. 
To address this limitation and better exploit benchmark data, we propose \textsc{SkillEval}, a new evaluation framework that performs cross-model, skill-level analysis. 
The framework integrates advances from two areas: 
(1) it leverages the meta-cognitive capabilities of LLMs to automatically discover interpretable skills from questions, and 
(2) it applies cognitive diagnosis models (CDMs) from psychometrics to estimate per-skill abilities of LLMs. 
We conduct experiments on 3,811 LLMs evaluated on 9,523 items across five benchmarks spanning diverse domains (MATH, BBH, GPQA, MuSR, IFEval).
Compared to the traditional average-based evaluation paradigm, \textsc{SkillEval} produces reliable, more calibrated and informative yet concise model profiles that more accurately predict model behavior on unseen items.
We also demonstrate the utility of \textsc{SkillEval} in two downstream tasks:
(1) LLM routing, which dynamically assigns each query to a capable but efficient model, reducing inference cost by over 60\% while achieving improved performance, and
(2) Computerized Adaptive Testing (CAT), where an unseen model can be efficiently evaluated using a trained CDM.
Furthermore, we show that these skill profiles provide useful insights, for example, how benchmarks share skills and how instruction-tuned models differ from their base models.
\end{abstract}

\section{Introduction}
\label{sec:intro}

Benchmarks for large language models (LLMs) have rapidly proliferated in recent years. While this creates new opportunities, it also raises a key challenge: how to effectively leverage a large collection of benchmarks for comprehensive LLM evaluation? 
Current evaluation practices remain quite raw. 
The common approach is to evaluate a model on all items in a benchmark and then report the average score. 
This naive statistical way collapses performance across items into a single number, which obscures fine-grained capability differences \citep{cui2026llmsexhibitcoherentknowledge}. 
For example, two LLMs with similar overall accuracy on a math benchmark (0.63 vs. 0.65) may exhibit very different profiles across sub-skills such as algebra and geometry.
This leads to a natural question: \emph{Given a set of test items, how can we go beyond a scalar score to a more informative, human-understandable capability estimate?}

To answer this, we develop a new evaluation framework that aims to assess the \emph{latent abilities} underlying observed item scores, and represents them as fine-grained skill profiles of LLMs. 
This problem has long been studied in psychometrics and the education community using various latent variable models. 
A representative family of approaches is Item Response Theory \citep[IRT;][]{lord2008statistical}, which jointly models test-taker traits (e.g., ability) and item traits (e.g., difficulty and discrimination). 
IRT models are also increasingly used in machine learning, but typically for data-centric tasks such as dataset analysis or distillation \citep{kipnis2025metabench, polo2024tinybenchmarks, zouhar-etal-2025-select, rodriguez-etal-2021-evaluation,cui2026makesharderexplainingquestion}, and less often for LLM evaluation as it still produces a scalar overall ability score that suffers from capability collapse. 
Some recent work explores multidimensional-IRT \citep{reckase200618} and its neural variants that learn ability vectors \citep{chen2025irtnet, cui-sachan-2025-investigating, song2025irtrouter, gor-etal-2024-great}. 
However, these multidimensional models are known to suffer from poor interpretability and non-identifiability, making them less suitable for reliable evaluation.

In this study, we propose \textsc{SkillEval}, a new evaluation framework that identifies latent skills and estimates LLM abilities at the skill level. 
The framework consists of a skill discovery module, which extracts shared skills from problems, and a neural cognitive diagnosis model  \citep[CDM;][]{wang2020neural}, which estimates per-skill abilities based on model performance on skill-associated items.
We evaluate \textsc{SkillEval} on a large-scale dataset of 36M model--item response records, covering 3,811 models and 9,523 items across five benchmarks from various domains, including MATH~\citep{hendrycks2021measuring}, BBH~\citep{suzgun-etal-2023-challenging}, GPQA~\citep{rein2024gpqa}, MuSR~\citep{sprague2024musr}, and IFEval~\citep{zhou2023instruction}.
We conduct extensive experiments and demonstrate the superiority of \textsc{SkillEval} from various aspects:
\begin{itemize}[leftmargin=15pt]
    \item Compared to MIRT-style methods, \textsc{SkillEval} extracts high-quality, human-understandable skills and produces stable ability estimates across runs, ensuring reproducible and robust evaluation.
    \item \textsc{SkillEval} accurately predicts model performance at both the item (with ${\rm AUC}=81.73$) and benchmark (with Pearson $r=0.989$) levels, outperforming strong baselines. In a downstream routing task where the goal is to assign queries to a cheap yet capable model, \textsc{SkillEval} matches the strongest LLM at $25\%$ of the cost ($\tau{=}0.80$, $99.9\%$ of strongest accuracy), and exceeds it by $3.7$ points at $37\%$ of the cost ($\tau{=}0.95$). It also profiles previously unseen LLMs efficiently without retraining the shared network: $20$ adaptively chosen answers yield a mean benchmark-score error comparable to $100$ randomly chosen answers.
    \item In addition, \textsc{SkillEval} serves as a useful tool for analyzing dataset structure and model behavior, such as skill coverage across benchmarks, co-mastery patterns among skills, and skill-level differences between base and instruction-tuned models, offering useful insights into data curation and model training. 
\end{itemize}



%
%
%
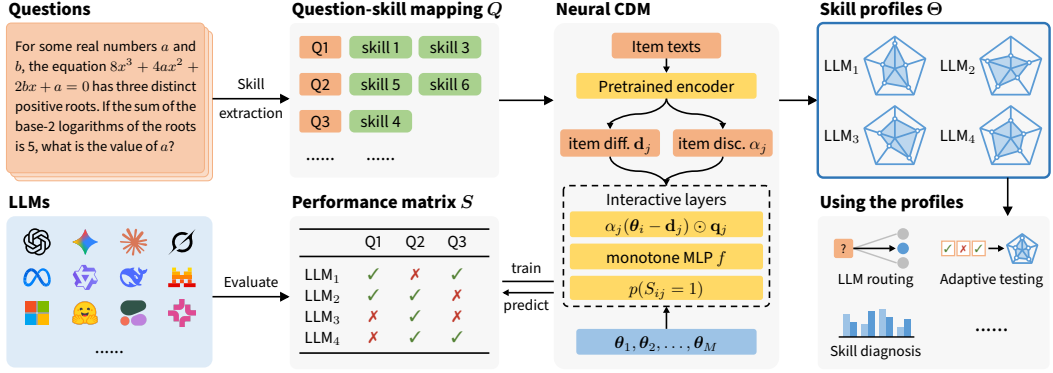
\begin{figure}[t]
\centering
\IfFileExists{T1SourceSansThree-LF.fd}{\def\seFontFamily{SourceSansThree-LF}}{%
  \IfFileExists{T1SourceSansPro-LF.fd}{\def\seFontFamily{SourceSansPro-LF}}{%
    \edef\seFontFamily{\sfdefault}}}
\def\seLogoDir{figures/llm_logos}

\definecolor{orange60}{HTML}{F8CBAD}
\definecolor{orange40}{HTML}{F4B183}
\definecolor{green40}{HTML}{A9D18E}
\definecolor{greenDark}{HTML}{548235}
\definecolor{gold40}{HTML}{FFD966}
\definecolor{blue80}{HTML}{DEEBF7}
\definecolor{blue60}{HTML}{BDD7EE}
\definecolor{blue40}{HTML}{9DC3E6}
\definecolor{blueBase}{HTML}{5B9BD5}
\definecolor{blueDark}{HTML}{2E75B6}
\definecolor{crossRed}{HTML}{C0392B}
\definecolor{panel}{HTML}{F7F7F7}
\definecolor{mutedgrey}{HTML}{BFBFBF}
\colorlet{questionbox}{orange60}   
\colorlet{salmon}{orange40}        
\colorlet{skillgreen}{green40}     
\colorlet{gold}{gold40}            
\colorlet{llmbox}{blue80}          
\colorlet{thetablue}{blue40}       
\colorlet{radarfill}{blue60}
\colorlet{radarline}{blueBase}
\colorlet{profileframe}{blueDark}
\colorlet{okgreen}{greenDark}      
\colorlet{wrongred}{crossRed}       

\newcommand\seFs[1]{\fontsize{#1bp}{#1bp}\selectfont}
\newcommand\seTitleFont{\seFs{11.5}\bfseries}   
\newcommand\seLabelFont{\seFs{10.5}}            
\newcommand\seSmallFont{\seFs{9.5}}             

\newcommand\seSconn[5][]{%
  \draw[#1] (#2,#3)
    .. controls (#2,{#3+(#5-#3)/4}) and ({(3*#2+#4)/4},{(#3+#5)/2})
    .. ({(#2+#4)/2},{(#3+#5)/2})
    .. controls ({(#2+3*#4)/4},{(#3+#5)/2}) and (#4,{#5-(#5-#3)/4})
    .. (#4,#5);}

\newcommand\seLogo[4]{\node[inner sep=0pt] at (#1,#2)
  {\includegraphics[width=#3bp, height=#3bp, keepaspectratio, draft=false]{\seLogoDir/#4}};}

\newcommand\seRadarR{20bp}                

\tikzset{
  lbl/.style  = {inner sep=0pt, anchor=base},        
  lblL/.style = {inner sep=0pt, anchor=base west},   
  arr/.style  = {line width=1bp, -{Triangle[length=6bp, width=6bp]}},
  arrS/.style = {line width=1bp, -{Triangle[length=4bp, width=4bp]}},
  pics/ok/.style    = {code={\node[inner sep=0pt, text=okgreen, font=\seFs{9.5}] {\ding{51}};}},
  pics/wrong/.style = {code={\node[inner sep=0pt, text=wrongred, font=\seFs{10}] {\ding{55}};}},
  pics/routingicon/.style = {code={%
      \foreach \y in {-9,9}
        \draw[mutedgrey, line width=0.9bp] (-14,0) -- (13,\y);
      \draw[arrS] (-14,0) -- (12.5,0);
      \fill[salmon, rounded corners=1.5bp] (-26,-6) rectangle (-14,6);
      \node[inner sep=0pt, font=\seFs{8}] at (-20,0) {?};
      \foreach \y/\c in {-9/mutedgrey, 0/radarline, 9/mutedgrey}
        \fill[\c] (17,\y) circle[radius=3.6bp];}},
  pics/adaptiveicon/.style = {code={%
      \foreach \x/\m/\c in {-27/51/okgreen, -17.5/55/wrongred, -8/51/okgreen} {
        \fill[white] (\x-4,-4.5) rectangle (\x+4,4.5);
        \draw[salmon, line width=0.8bp] (\x-4,-4.5) rectangle (\x+4,4.5);
        \node[inner sep=0pt, text=\c, font=\seFs{6.5}] at (\x,0) {\ding{\m}};
      }
      \draw[arrS] (-2,0) -- (8,0);
      \begin{scope}[shift={(19,1)}]
        \def\seRadarR{10bp}
        \pic {radar=0.75/0.45/0.60/0.70/0.40};
      \end{scope}}},
  pics/diagnosisicon/.style = {code={%
      \foreach \i/\a/\b in {0/14/9, 1/8/16, 2/17/12, 3/6/11} {
        \fill[thetablue] ({-20+11*\i},10) rectangle ++(4.2,-\a);
        \fill[radarline] ({-15.6+11*\i},10) rectangle ++(4.2,-\b);
      }
      \draw[line width=0.9bp] (-23,10) -- (24,10);}},
  pics/radar/.style args = {#1/#2/#3/#4/#5}{code={%
      \fill[radarfill] (90:{#1*\seRadarR}) -- (162:{#2*\seRadarR}) -- (234:{#3*\seRadarR})
                       -- (306:{#4*\seRadarR}) -- (18:{#5*\seRadarR}) -- cycle;
      \begin{scope}[radarline, line width=1.1bp, line join=round]
        \foreach \a in {90,162,234,306,18} \draw (0,0) -- (\a:\seRadarR);
        \draw (90:\seRadarR) -- (162:\seRadarR) -- (234:\seRadarR) -- (306:\seRadarR)
              -- (18:\seRadarR) -- cycle;
        \draw (90:{#1*\seRadarR}) -- (162:{#2*\seRadarR}) -- (234:{#3*\seRadarR})
              -- (306:{#4*\seRadarR}) -- (18:{#5*\seRadarR}) -- cycle;
      \end{scope}
      \foreach \a/\v in {90/#1,162/#2,234/#3,306/#4,18/#5}
        \filldraw[fill=white, draw=radarline, line width=0.9bp]
          (\a:{\v*\seRadarR}) circle[radius=1.2bp];}},
}

\resizebox{\linewidth}{!}{\fontfamily{\seFontFamily}\selectfont%
\begin{tikzpicture}[x=1bp, y=-1bp]
\useasboundingbox (0,0) rectangle (656,234);


\begin{scope}[shift={(4,0)}]
  \node[lblL, font=\seTitleFont] at (1,11) {Questions};
  \foreach \d/\tint in {5/45, 2.5/70}
    \filldraw[fill=questionbox!\tint!white, draw=salmon, line width=0.6bp,
              rounded corners=3bp] (\d,16+\d) rectangle (123+\d,109+\d);
  \filldraw[fill=questionbox, draw=salmon, line width=0.6bp, rounded corners=3bp]
    (0,16) rectangle (123,109);
  \node[lblL] at (5.25,33.6) {\parbox[t]{112.5bp}{%
      \fontsize{9bp}{12.4bp}\selectfont\frenchspacing
      \tolerance=10000\hbadness=10000
      For some real numbers $a$ and $b$, the equation $8x^3+4ax^2+2bx+a=0$
      has three distinct positive roots. If the sum of the base\nobreakdash-2
      logarithms of the roots is 5, what is the value of~$a$?}};
\end{scope}

\draw[arr] (132,63) -- (180,63);
\node[lbl, font=\seSmallFont] at (156,57) {Skill};
\node[lbl, font=\seSmallFont] at (156,75) {extraction};

\begin{scope}[shift={(180,0)}]
  \node[lblL, font=\seTitleFont] at (1,11) {Question-skill mapping $\bm{Q}$};
  \fill[panel, rounded corners=3.7bp] (0,16) rectangle (130,112);
  \begin{scope}[font=\seLabelFont]
    \foreach \q/\y in {1/24, 2/47, 3/70} {
      \fill[salmon, rounded corners=1.6bp] (6,\y) rectangle (32,\y+14);
      \node[lbl] at (19,\y+10.5) {Q\q};
    }
    \foreach \x/\y/\s in {37/24/1, 80/24/3,
                          37/47/5, 80/47/6,
                          37/70/4} {
      \fill[skillgreen, rounded corners=3.2bp] (\x,\y) rectangle (\x+39,\y+14);
      \node[lbl] at (\x+19.5,\y+10.5) {skill \s};
    }
  \end{scope}
  \node[lbl, font=\seFs{12}] at (19,101) {......};
  \node[lbl, font=\seFs{12}] at (56.5,101) {......};
\end{scope}

\draw[arr] (310,64) -- (344,64);

\begin{scope}[shift={(4,0)}]
  \node[lblL, font=\seTitleFont] at (1,131) {LLMs};
  \fill[llmbox, rounded corners=5.3bp] (0,136) rectangle (128,230);
  \foreach \f [count=\k from 0] in {openai.pdf, gemini.pdf, claude.pdf, grok.pdf,
                                     meta.pdf, qwen.pdf, deepseek.pdf, mistral.pdf,
                                     microsoft.png, huggingface.png, cohere.png, ai2.png} {
    \pgfmathsetmacro\cx{19+30*mod(\k,4)}
    \pgfmathsetmacro\cy{152+22*floor(\k/4)}
    \ifnum\k=8 \seLogo{\cx}{\cy}{14}{\f}\else \seLogo{\cx}{\cy}{17}{\f}\fi
  }
  \node[lbl, font=\seFs{12}] at (64,221) {......};
\end{scope}

\draw[arr] (132,186) -- (180,186);
\node[lbl, font=\seSmallFont] at (156,180) {Evaluate};

\begin{scope}[shift={(180,0)}]
  \node[lblL, font=\seTitleFont] at (1,131) {Performance matrix $\bm{S}$};
  \fill[panel, rounded corners=4.4bp] (0,136) rectangle (130,231);
  \foreach \y in {144, 161, 221} \draw[line width=1bp] (6,\y) -- (124,\y);
  \begin{scope}[font=\seLabelFont]
    \foreach \q/\x in {1/52, 2/78, 3/104} \node[lbl] at (\x,155) {Q\q};
    \foreach \m/\y in {1/172, 2/185, 3/198, 4/211}
      \node[lblL] at (9,\y+3.6) {LLM$_{\m}$};
  \end{scope}
  \pic at (52,172) {ok};    \pic at (78,172) {wrong}; \pic at (104,172) {ok};
  \pic at (52,185) {ok};    \pic at (78,185) {ok};    \pic at (104,185) {wrong};
  \pic at (52,198) {wrong}; \pic at (78,198) {ok};    \pic at (104,198) {wrong};
  \pic at (52,211) {wrong}; \pic at (78,211) {ok};    \pic at (104,211) {ok};
\end{scope}

\draw[arrS] (312,176.5) -- (353,176.5);
\draw[arrS] (353,183.5) -- (312,183.5);
\node[lbl, font=\seSmallFont] at (327,171.5) {train};
\node[lbl, font=\seSmallFont] at (327,194) {predict};

\begin{scope}[shift={(344,0)}]
  \node[lblL, font=\seTitleFont] at (1,11) {Neural CDM};
  \fill[panel, rounded corners=7bp] (0,16) rectangle (140,231);

  \begin{scope}[font=\seLabelFont]
    \fill[salmon, rounded corners=2.2bp] (36,22) rectangle (104,38);
    \node[lbl] at (70,34) {Item texts};
    \draw[arrS] (70,38) -- (70,47);
    \fill[gold, rounded corners=1.9bp] (14,47) rectangle (126,63);
    \node[lbl] at (70,59) {Pretrained encoder};
    \seSconn[arrS]{70}{63}{35}{82}
    \seSconn[arrS]{70}{63}{105}{82}
    \fill[salmon, rounded corners=2.3bp] (4,82) rectangle (66,99);
    \node[lbl] at (35,94.5) {item diff.\ $\mathbf{d}_j$};
    \fill[salmon, rounded corners=2.3bp] (74,82) rectangle (136,99);
    \node[lbl] at (105,94.5) {item disc.\ $\alpha_j$};
    \seSconn[line width=1bp]{35}{99}{70}{117}
    \seSconn[arrS]{105}{99}{70}{117}

    \draw[line width=1bp, dash pattern=on 4bp off 3bp, rounded corners=2.8bp]
      (6,117) rectangle (134,192);
    \node[lbl] at (70,129) {Interactive layers};
    \fill[gold, rounded corners=1.7bp] (10,133) rectangle (130,149);
    \node[lbl] at (70,145) {$\alpha_j(\bm{\theta}_i-\mathbf{d}_j)\odot\mathbf{q}_j$};
    \fill[gold, rounded corners=1.7bp] (10,153) rectangle (130,169);
    \node[lbl] at (70,165) {monotone MLP $f$};
    \fill[gold, rounded corners=1.7bp] (10,173) rectangle (130,188);
    \node[lbl] at (70,184.5) {$p(S_{ij}=1)$};

    \draw[arrS] (70,206) -- (70,192);
    \fill[thetablue, rounded corners=1.9bp] (14,206) rectangle (126,222);
    \node[lbl] at (70,218) {$\bm{\theta}_1, \bm{\theta}_2, \ldots, \bm{\theta}_M$};
  \end{scope}
\end{scope}

\draw[arr] (484,64) -- (508,64);

\begin{scope}[shift={(508,0)}]
  \node[lblL, font=\seTitleFont] at (1,11) {Skill profiles $\bm{\Theta}$};
  \filldraw[fill=panel, draw=profileframe, line width=1.5bp, rounded corners=3.6bp]
    (0,16) rectangle (144,112);
  \foreach \m/\x/\y in {1/3.6/45.8, 2/75.6/45.8, 3/3.6/89, 4/75.6/89}
    \node[lblL, font=\seLabelFont] at (\x,\y) {LLM$_{\m}$};
  \pic at (47.9,44.3)  {radar=0.78/0.40/0.55/0.72/0.35};   
  \pic at (119.9,44.3) {radar=0.40/0.76/0.68/0.35/0.60};   
  \pic at (47.9,87.5)  {radar=0.60/0.50/0.30/0.76/0.74};   
  \pic at (119.9,87.5) {radar=0.55/0.62/0.74/0.48/0.42};   

  \draw[arr] (118,112) -- (118,136);
  \node[lblL, font=\seTitleFont] at (1,131) {Using the profiles};
  \fill[panel, rounded corners=3.7bp] (0,136) rectangle (144,231);
  \pic at (36,156)  {routingicon};
  \node[lbl, font=\seSmallFont] at (36,178) {LLM routing};
  \pic at (108,156) {adaptiveicon};
  \node[lbl, font=\seSmallFont] at (108,178) {Adaptive testing};
  \pic at (36,201)  {diagnosisicon};
  \node[lbl, font=\seSmallFont] at (36,223) {Skill diagnosis};
  \node[lbl, font=\seFs{14}] at (108,207) {......};
\end{scope}

\end{tikzpicture}}
\caption{We first extract skills from the questions and construct a question-to-skill mapping (i.e., the $Q$-matrix). Then, using this $Q$-matrix together with the performance matrix of models on items, we train a neural CDM to estimate each model's skill-level abilities $\theta$.}
\label{fig:main}
\end{figure}

\section{Profiling LLMs' Skills with Cognitive Diagnostic Models}
\label{sec:method}

\subsection{Problem Statement}
Given a set of models to be evaluated $\{\mathcal{M}_1, ..., \mathcal{M}_M \}$ and a collection of test items $\{x_1, ..., x_N\}$ from one or multiple benchmarks, we represent their performance using a matrix $\mathcal{S} \in \mathbb{R}^{M \times N}$, where each entry $\mathcal{S}_{ij}$ denotes the score of the $i$-th model on the $j$-th item. 
We formulate LLM profiling as a function $\mathcal{F}$ that takes $\mathcal{S}$ as input and outputs some estimates for each model's capabilities, i.e., $\mathcal{F}(\mathcal{S}) \rightarrow \{\theta_1, ..., \theta_M\}$, where $\theta_i$ is a quantitative measurement of the capabilities of model $i$.

The common practice in evaluation is to model $\mathcal{F}$ as a simple average over individual items, i.e., $\theta_i = \frac{1}{n} \sum_{j=1}^{n} S_{ij}$.
While this provides a concise summary of model performance, it suffers from two key limitations. First, it compresses the model's performance into a single scalar metric, failing to capture variations across fine-grained skills. 
Second, the naive average can be affected by bias induced by the difficulty distribution of the test items, and may not faithfully reflect the model's true ability. Indeed, this type of ability modeling assumes that all items in a given benchmark contribute equally to the estimation of a single ability dimension, which is typically an implausible assumption. 

To address these limitations, we propose a new paradigm for profiling LLMs. 
We assume that the items in the benchmark are associated with a set of underlying latent skills $\mathcal{C} = \{c_1, \ldots, c_K\}$.
Inspired by psychometric approaches for assessing latent traits, we use Cognitive Diagnosis Models (CDMs) to construct the profiling function $\mathcal{F}_{\rm CD}(\mathcal{S}) \rightarrow \Theta \in \mathbb{R}^{M \times K}$, where the profile $\theta_i$ for each model is a $k$-dimensional vector representing its proficiency over the set of skills.
An overview of our method is shown in Figure \ref{fig:main}.

\subsection{Q-Matrix Construction}
\label{sec:method:qmatrix}
We use a binary Q-matrix $\mathcal{Q} \in \{0, 1\}^{N \times K}$ to denote the mapping between $N$ items and $K$ skills, where each entry $\mathcal{Q}_{ij}=1$ indicates the $i$-th item is associated with the $j$-th skill and 0 otherwise.
Manually defining and annotating skills is challenging, costly, and difficult to scale.
Recent work shows that LLMs can automatically identify meaningful skills from problem content \citep{didolkar2024metacognitive,he2026stat}. 
We adapt this approach to our setting, consisting of two main steps: skill labeling and skill clustering.

\paragraph{Skill labeling.}
For each item $j$, we prompt \texttt{GPT-4o-mini} with the question stem and ask
for (i)~a single \emph{primary} skill and (ii)~the full list of skills
required for the solution (exact prompt in Appendix~\ref{app:qmatrix}). 
We use the full list rather than only the primary skill: primary-skill assignments produce Q-matrix rows
of mean row-sum $1.0$, which under-constrains NCDM. 
Full-list extraction yields a mean of $2.9$ skills per item and $13{,}276$ unique skill strings
across $9{,}523$ items.

\paragraph{Skill clustering.}
Because skills are extracted independently from different problems, they may contain redundancy and noise. 
To obtain a more compact and coherent skill set, we cluster similar skills and merge them accordingly. 
\cite{didolkar2024metacognitive} employ an LLM to perform this grouping process; however, such approaches may not scale well to a large number of skills. Therefore, we adopt an embedding-based clustering method, which is more efficient and scalable.
Each unique skill string is first encoded by the pretrained SBERT
\texttt{all-mpnet-base-v2} \citep{reimers-gurevych-2019-sentence} into
$\mathbb{R}^{768}$. 
Then, we use the hierarchical agglomerative clustering (HAC) algorithm with cosine distance and average linkage, specifying the number of final clusters as $K$.

\paragraph{Q-matrix assembly.}
We set $\mathcal{Q}_{jk}=1$ iff at least one of item $j$'s extracted skills falls
into cluster $k$. This multi-skill assignment reduces the fraction of
clusters with fewer than ten items from $43\%$ under primary-skill assignment to $36\%$ at
$K{=}100$\footnote{We adopt $K{=}100$ from $\{50,100,150,200,300,400,500\}$ which achieves a better balance between interpretability and performance. See Appendix~\ref{app:ksweep} for more details.}, yielding a mean column-sum of ${\sim}\,139$ items per skill
(mean row-sum $1.5$ clusters per item, reflecting the compression of the
$2.9$ extracted skill strings into shared clusters), the regime in which
NCDM trains stably.

\subsection{Neural CDM}
\label{sec:method:ncdm}
We adopt the Neural Cognitive Diagnosis Model
\citep{wang2020neural}. The probability that LLM $s$ answers item $j$ correctly
is estimated as:
\begin{equation}
    p(\mathcal{S}_{ij}=1) \;=\; {\rm sigmoid}\!\Big(f\!\big(\, \alpha_{j}\,
    ({\boldsymbol{\theta}}_i - \boldsymbol{d}_{j}) \odot \boldsymbol{q}_j\,\big)\Big),
    \label{eq:ncdm}
\end{equation}
where $\boldsymbol{\theta}_i \in [0,1]^K$ represents the learned skill mastery of model $\mathcal{M}_i$, $\alpha_j \in [0,1]$ is the discrimination parameter of item $j$, $\boldsymbol{d}_j \in [0,1]^K$ is its per-skill difficulty vector, and $\boldsymbol{q}_j \in \{0,1\}^K$ is the corresponding Q-matrix row. 
The scalar $\alpha_j$ scales all skill dimensions equally. The function $f$ is a three-layer MLP with ${\rm PosLinear}$ layers. Each ${\rm PosLinear}$ layer uses non-negative weights, $\mathbf{W}^{+}=\max(\mathbf{W},0)$. 
Together with the non-decreasing sigmoid activations, this ensures that increasing any mastery component cannot decrease the predicted probability, holding all other inputs fixed.

\subsection{Learning Item Parameters from Text}
\label{sec:method:text}
The original NCDM treats item discrimination $\alpha_{j}$  and difficulty $\boldsymbol{d}_{j}$ as model parameters learned during training, preventing predictions on unseen items. 
To overcome this limitation, we follow previous context-aware approaches \citep{gor-etal-2024-great, cui-sachan-2025-investigating} to use a lightweight neural network to learn item parameters based on their text features. 
Specifically, for a given item $x_j$, we first use a pretrained encoder to obtain its embedding $\boldsymbol{e}_j$. 
Then, we compute $\alpha_{j}$ and $\boldsymbol{d}_{j}$ via:
\begin{equation}
    \alpha_{j} = \sigma\!\big(\mathbf{w}_\alpha^\top \boldsymbol{e}_j + b_\alpha \big),
    \qquad
    \boldsymbol{d}_{j} = \sigma\!\big(\mathbf{W}_d\,\boldsymbol{e}_j + {b}_d\big).
    \label{eq:text}
\end{equation}
To maintain computational efficiency, we freeze the pretrained encoder and only optimize the projection parameters $\mathbf{w}_{\alpha}, \mathbf{W}_{d}$ and $b_{\alpha}, {b}_{d}$.

\section{Evaluation of Skill Profiles}
\label{sec:experiments}

\paragraph{Experiment Setup.}
We train our text-conditioned NCDM on a subset of the RouterEval dataset~\citep{huang2025routereval}, including 3,811 LLMs evaluated on 9,523 items from five benchmarks: MATH, BBH, GPQA, MuSR, and IFEval.
For the number of skills, we set $K{=}100$ as the main experimental configuration, as it achieves a better balance between interpretability and performance. We provide comparisons with other values of $K$ in Appendix~\ref{app:ksweep}.

\subsection{Quality of Discovered Skills} \label{sec:eval_skill_quality}
Since our evaluation is centered around skills, it is crucial to assess the quality of these LLM-discovered skills. 
We consider the following three criteria:
(1) \textbf{Semantic clarity}, which measures whether a skill description is clear, unambiguous, and self-contained;
(2) \textbf{Level of abstraction}, which captures the granularity of a skill, where a good skill should be neither overly broad nor overly specific to particular questions;
(3) \textbf{Diversity}, which evaluates whether different skills are non-redundant and capture distinct aspects.
For the first two criteria, we adopt LLM-as-a-judge. 
For the last one, we use intra-list diversity defined as averaged pair-wise dissimilarities, which is widely used in recommendation systems.

\begin{wraptable}[10]{r}{0.42\textwidth}
\vspace{-7mm}
\caption{Comparison between the LLM-extracted skill set (${\rm K}=100$) and an expert-defined mathematics skill set (the Common Core standards; ${\rm K}=406$).}\label{tab:skill_quality}
\centering
\small
\setlength\tabcolsep{2pt}
\begin{tabular}{lll}
\toprule
            & \textbf{LLM-extracted} & \textbf{Expert-defined} \\ \midrule
Clarity     & 4.07 \includegraphics[height=3mm]{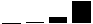}          & 4.14 \includegraphics[height=3mm]{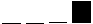}            \\
Granularity & 2.26 \includegraphics[height=3mm]{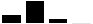}          & 2.72 \includegraphics[height=3mm]{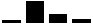}           \\
Diversity   & 0.83 \includegraphics[height=3mm]{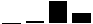}               & 0.76 \includegraphics[height=3mm]{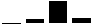}                \\ \bottomrule
\end{tabular}
\end{wraptable}
The experimental results are shown in Table \ref{tab:skill_quality}. 
As a reference, we compare our skills with the Common Core State Standards\footnote{\url{https://corestandards.org/}}, a widely used expert-curated framework for K--12 math education. 
Our skills achieve a similar level of semantic clarity to the Common Core. 
In terms of granularity, they are slightly more fine-grained. 
They also show lower variance (${\rm std}$. 0.29 vs. 0.55), indicating more consistent granularity across skills, which is desirable. 
In addition, our skills are more diverse (${\rm avg}$. 0.83 vs. 0.76), likely because they cover more domains and reduce redundancy through clustering.
Overall, our automatically extracted skills are competitive with manually designed ones along key dimensions.

\paragraph{Human validation of skill--question assignments.}
Clear skill descriptions do not by themselves establish that the skills are relevant to their assigned questions. We therefore evaluate the skill--question assignments through human annotation. Three annotators, each holding an MSc in engineering, evaluated $780$ skill--question pairs covering all $100$ skills. We sampled up to $10$ questions uniformly at random from each skill's assigned items, retaining all items for smaller clusters. Annotators were shown revised descriptive skill names and question text, and judged whether answering each question required the named skill. Majority voting accepted $652$ of $780$ assignments ($83.6\%$), with an acceptance rate of $81.7\%$ when averaged equally across skills. Inter-annotator agreement was $0.715$ in mean pairwise Cohen's kappa and $0.714$ in Krippendorff's nominal alpha; both measures equal $1$ for perfect agreement and $0$ for agreement expected by chance. These judgments provide evidence for the semantic relevance of the sampled assignments.

\subsection{Stability}\label{sec:stability}
For an evaluation method, the stability of its results is essential.
However, multidimensional latent ability models commonly suffer from non-identifiability: the $\theta$ vectors learned in repeated training runs are inconsistent, as they are affected by rotation, permutation, and random initialization.
This prevents such models from serving as a reliable and reproducible means of evaluation.
\textsc{SkillEval} instead ties each dimension of $\theta$ to a named skill through a fixed Q-matrix, which anchors the coordinate system across runs.

We test whether ability profiles remain consistent across retraining, and compare \textsc{SkillEval} with IRTNet~\citep{chen2025irtnet}, a high-dimensional neural IRT model that represents each LLM with a $d$-dimensional latent ability vector.
We train \textsc{SkillEval} with five random seeds and IRTNet with different dimensionalities.
We compute Pearson correlations across learned $\mathbf{\Theta}$ across different runs in Table \ref{tab:theta_stability}.
Averaged over dimensions and run pairs, the correlation is $0.96$ for \textsc{SkillEval} and $\approx0.00$ for each IRTNet configuration. 
To rule out differences that arise merely from different coordinate systems, we also compare runs after Procrustes alignment, which finds the rotation that best maps one run's $\theta$ vectors onto another's.
Despite this, IRTNet's mean correlations rise only to $0.10$--$0.43$.
Thus, \textsc{SkillEval} supports more reliable and consistent evaluation of LLMs.

\begin{table}[t]
\small
\centering
\caption{Mean Pearson correlations between ability matrices $\Theta$ across run pairs, with standard deviations across pairs in parentheses. \textbf{Flat} correlates all entries of the two ability matrices after arranging them into vectors. \textbf{Per-LLM} correlates each LLM's ability profile between runs and averages over LLMs. These summaries can
be high when some dimensions consistently take larger values than others. \textbf{Per-dim} instead compares values across LLMs separately for each dimension, then averages the correlations over dimensions.  It therefore tests whether each dimension assigns similar relative values to LLMs across runs. \textbf{Raw} uses the dimensions as learned. \textbf{Orthogonal} applies the rotation or reflection that best aligns one run's ability matrix with the other \citep{schonemann1966}, and \textbf{Full} additionally centres and rescales the matrices before alignment \citep{gower1975}.} \label{tab:theta_stability}
\begin{tabular}{lccccc}
\toprule
\multirow{2}{*}{\textbf{Model}} & \multirow{2}{*}{\textbf{Flat}} & \multirow{2}{*}{\textbf{Per-LLM}} & \multicolumn{3}{c}{\textbf{Per-dim}}               \\
                                &                                &                                   & \textbf{Raw} & \textbf{Orthogonal} & \textbf{Full} \\ \midrule
IRTNet ($d{=}232$) & 0.02 \std{0.06} & 0.02 \std{0.06} & 0.00 \std{0.01} & 0.43 \std{0.22} & 0.43 \std{0.21} \\
IRTNet ($d{=}512$) & 0.00 \std{0.04} & 0.00 \std{0.04} & 0.00 \std{0.00} & 0.17 \std{0.02} & 0.17 \std{0.01} \\
IRTNet ($d{=}1024$) & 0.01 \std{0.00} & 0.01 \std{0.00} & 0.00 \std{0.00} & 0.10 \std{0.02} & 0.10 \std{0.02} \\
\textsc{SkillEval} (${\rm K}=100$) & \textbf{0.99} \std{0.00} & \textbf{0.99} \std{0.01} & \textbf{0.96} \std{0.00} & \textbf{0.96} \std{0.00} & \textbf{0.96} \std{0.00} \\ \bottomrule
\end{tabular}
\end{table}

\subsection{Predictive Validity} \label{sec:eval_theta}
Finally, we evaluate how well $\Theta$ predicts model behavior in independent tasks. 
On held-out items, the mean per-LLM correlation between $\Theta$ and per-skill accuracy is $0.73$ across five seeds,\footnote{Each skill is weighted by its number of held-out items. Appendix~\ref{app:validity_controls} reports controls for overall, benchmark-level, and subtask-level accuracy.} validating $\Theta$ as a measurement target. We further test the full predictive power of $\Theta$ via item- and benchmark-level prediction with the trained NCDM.

\paragraph{Item-level Prediction.}
 
We consider both in-distribution (ID) and out-of-distribution (OOD) settings. 
In the ID setting, we perform an (LLM $\times$ Item) interaction-wise split, where all items are observed during training. 
In the OOD setting, we use an item-wise split, holding out a subset of items that are never seen during training.

\begin{wraptable}{r}{0.5\textwidth}
\vspace{-7mm}
\centering
\small
\caption{AUC and accuracy (Acc) under two settings: ID (interaction-wise split) and OOD (item-wise split). For Acc, a response is predicted correct when its predicted probability is at least $0.5$. IRT 2PL has no item features and therefore cannot generalise to held-out items; its OOD column is left blank. OOD values are means over seeds, five for \textsc{SkillEval} and three for IRTNet. ID values are single runs.} \label{tab:item_prediction}
\setlength\tabcolsep{3pt}
\begin{tabular}{lcccc}
\toprule
\multirow{2}{*}{\textbf{Model}} & \multicolumn{2}{c}{\textbf{AUC}} & \multicolumn{2}{c}{\textbf{Acc}} \\
 & \textbf{ID} & \textbf{OOD} & \textbf{ID} & \textbf{OOD} \\ \midrule
IRT (2PL) & \textbf{86.16} & -- & \textbf{78.76} & -- \\
\textsc{IRTNet} ($d{=}232$) & 81.38 & 75.04 & 74.21 & 69.05 \\
\textsc{IRTNet} ($d{=}512$) & 81.36 & 75.14 & 74.21 & 69.07 \\
\textsc{IRTNet} ($d{=}1024$) & 81.74 & \textbf{75.35} & 74.68 & \textbf{69.38} \\ \midrule
\textsc{SkillEval} (${\rm K}=100$) & 81.73 & 72.15 & 75.57 & 67.62 \\ \bottomrule
\vspace{-5mm}
\end{tabular}
\end{wraptable}
We compare \textsc{SkillEval} with 2PL IRT and IRTNet in Table~\ref{tab:item_prediction}.
The 2PL model scores highest in the ID setting, as it learns a separate difficulty and discrimination for every item, but for the same reason it cannot score items unseen during training.
In the ID setting, \textsc{SkillEval} performs on par with IRTNet, even though IRTNet has $34$ to $39$ times as many trainable parameters: $22.0$M to $25.2$M across the three sizes, against $0.64$M for \textsc{SkillEval}.
In the OOD setting, IRTNet is ahead by $3.2$ AUC and $1.8$ accuracy points.
This gap is expected: IRTNet lets every item use all of its latent dimensions, whereas \textsc{SkillEval} restricts each question to the dimensions of its assigned skills.
The same restriction is what gives each dimension of $\Theta$ a clear meaning and keeps it stable across runs, which IRTNet's dimensions lack (\S\ref{sec:stability}).
\textsc{SkillEval} thus remains competitive in prediction while providing a far more stable and interpretable ability profile.

\begin{figure}[h]
\centering
\includegraphics[width=\linewidth]{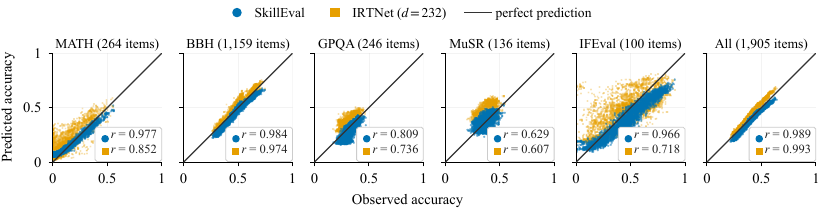}
\caption{Predicted versus observed benchmark accuracy on held-out items in the OOD setting. Each point represents one LLM. The horizontal axis shows its observed accuracy on the benchmark's held-out items; the vertical axis shows the mean predicted probability of a correct answer on those items. The diagonal indicates exact agreement, and $r$ is the Pearson correlation across $3{,}811$ LLMs. IRTNet uses $d{=}232$.} 
\label{fig:benchpred}
\end{figure}

\paragraph{Benchmark-level prediction.}
Benchmark accuracy is the quantity most often used to compare LLMs, so we ask whether item-level predictions recover each LLM's score on each benchmark.
In the OOD setting, we average an LLM's predicted probabilities over a benchmark's held-out items and compare the result with its observed accuracy on those items (Figure~\ref{fig:benchpred}).
Although the two methods reach similar correlations on the pooled score ($r{=}0.989$ vs.\ $0.993$), \textsc{SkillEval} is more accurate within every benchmark, most clearly on MATH ($0.977$ vs.\ $0.852$) and IFEval ($0.966$ vs.\ $0.718$).
\textsc{SkillEval}'s predictions lie in a narrow band along the diagonal, whereas IRTNet's spread widely: LLMs with nearly identical observed accuracy can receive very different predicted scores.
IRTNet also tends to overestimate accuracy, with most points above the diagonal on MATH, GPQA and MuSR, while \textsc{SkillEval} stays better calibrated.
This is because that IRTNet's per-benchmark errors for the same LLM point in different directions and largely cancel when averaged into the pooled score.
In addition, the pooled score is also dominated by BBH, which contributes 61\% of the held-out items and on which both methods already agree closely.
A good pooled correlation can therefore hide inaccurate per-benchmark estimates, and the per-benchmark comparison is the one that matters when LLMs are compared on a given benchmark.

\paragraph{Item-side versus model-side accuracy.}
Taken together, Table~\ref{tab:item_prediction} and Figure~\ref{fig:benchpred} reveal a trade-off between predicting items and comparing models.
IRTNet lets every item draw on all $d$ latent dimensions, so it can fit item-specific idiosyncrasies and attains higher pooled AUC on unseen items (+3.2 points).
\textsc{SkillEval} restricts each item to the dimensions of its assigned skills and summarises its difficulty along those skills with a single discrimination scalar, which limits how finely it can model individual items.
The same restriction forces all items that share a skill to constrain the same ability dimension, so each dimension of ${\Theta}$ is estimated from many related items.
As a result, \textsc{SkillEval} estimates where each LLM stands within a domain more accurately: it ranks LLMs better on every benchmark (Figure~\ref{fig:benchpred}) and yields ability profiles that are stable across retraining (Table~\ref{tab:theta_stability}).
The cost lies on the item side: difficulty and discrimination learned from text are less precise, and errors in the Q-matrix propagate directly into predictions.
For an evaluation framework, whose purpose is to compare models rather than to predict individual responses, we consider this trade-off favourable.

\section{Applications}

\subsection{Routing LLMs via Ability Profiles}
\label{sec:exp:routing}

LLM routing selects a model for each incoming query, with the aim of balancing answer accuracy and inference cost \citep{hu2024routerbench, huang2025routereval}. We first test whether skill profiles improve accuracy by selecting an LLM for each query, rather than always using the strongest model overall. For a test item $j$ \textsc{SkillEval} estimates the probability $P_i(j)=p(\mathcal{S}_{ij}{=}1)$ that LLM $i$ answers it correctly (Equation~\ref{eq:ncdm}). The router selects the LLM with the highest predicted probability: $
\hat{\imath}(j) \;=\; \arg\max_{i\in\{1,\dots,M\}} P_i(j).
\label{eq:route} $
We evaluate on the $N_{\text{test}}{=}1{,}905$ held out items of the OOD split with $M{=}3{,}811$ candidate LLMs. For routers that rank multiple candidates, $\mathrm{Acc@}k$  is the fraction of test items for which at least one of the top-$k$ LLMs answer correctly. Thus, $\mathrm{Acc@}1$ measures the accuracy of the selected LLM, while larger values of $k$ measure whether a shortlist contains a correct answer. 
\begin{wraptable}{r}{0.52\textwidth}
\vspace{-1mm}
\centering
\small
\caption{Routing performance on $1{,}905$ held-out items with $3{,}811$ candidate LLMs. All scores are percentages. Rows marked ``3 seeds'' report the mean over three trainings with different seeds, with the standard deviation in parentheses; \mbox{IRTNet} uses $d{=}1024$, its best size in Table~\ref{tab:item_prediction}. The strongest single LLM and KNN involve no training or randomness, Random is a single random draw, and the other rows are single training runs. Best results are \textbf{bold}.}
\label{tab:main}
\setlength\tabcolsep{3pt}
\resizebox{\linewidth}{!}{%
\begin{tabular}{lcccc}
\toprule
\textbf{Method} & \textbf{Acc@1} & \textbf{Acc@3} & \textbf{Acc@5} & \textbf{Acc@10} \\
\midrule
Random & 40.47 & -- & -- & -- \\
Strongest single LLM & 61.57 & -- & -- & -- \\
IRT 2PL & 61.10 & 77.74 & 81.94 & 85.93 \\
KNN & 62.05 & \textbf{78.01} & \textbf{83.10} & \textbf{88.98} \\ IRTNet, 3 seeds & 63.08 & 77.31 & 81.21 & 86.44 \\  & \std{0.98} & \std{0.55} & \std{1.60} & \std{0.76} \\
\midrule
\textsc{SkillEval} (${\rm K}{=}100$) & \textbf{65.67} & 77.95 & 82.94 & 87.51 \\
\textsc{SkillEval}, 3 seeds & 65.46 & 77.55 & 82.43 & 87.49 \\
 & \std{0.34} & \std{1.26} & \std{0.37} & \std{0.47} \\
\bottomrule
\end{tabular}%
}
\vspace{-3mm}
\end{wraptable}

\textbf{Baselines.} We compare against five baselines. \emph{Random} routing selects an LLM uniformly at random for each item. The \emph{strongest single LLM} always uses the model with the highest accuracy on the training items, \texttt{calme-3.2-instruct-78b}.\footnote{\url{https://huggingface.co/MaziyarPanahi/calme-3.2-instruct-78b}} The \emph{IRT~2PL} router ranks LLMs by the single ability value that a fitted 2PL model assigns to each of them \citep{lord2008statistical}, giving the same ranking for every query. Related work, IRT-Router \citep{song2025irtrouter}, instead uses multidimensional IRT for routing. The \emph{KNN} router \citep{chen2025irtnet} retrieves the $10$ training items with the most similar text embeddings to item $j$. For each candidate LLM, it estimates the probability of a correct answer as the fraction of these retrieved items that the LLM answered correctly. \emph{IRTNet} routes like \textsc{SkillEval}, selecting the LLM with the highest probability predicted by its own model.

Among the methods in Table~\ref{tab:main}, \textsc{SkillEval} achieves the highest $\mathrm{Acc@}1$: $65.67\%$, compared with $61.57\%$ for the strongest single LLM, $61.10\%$ for the IRT~2PL router, $62.05\%$ for KNN, and $63.08\%$ for IRTNet over three runs. Across three retrainings, \textsc{SkillEval}'s $\mathrm{Acc@}1$ is $(65.46 \pm 0.34)\%$ (mean $\pm$ standard deviation). IRTNet predicts individual responses more accurately (Table~\ref{tab:item_prediction}) but picks a correct LLM less often, since routing depends only on the top-ranked LLM for each question. For larger shortlists, KNN is competitive: its $\mathrm{Acc@}3$ is similar to that of \textsc{SkillEval}, and it achieves higher $\mathrm{Acc@}5$ and $\mathrm{Acc@}10$. The per-benchmark breakdown (Appendix~\ref{app:perbench}, Table~\ref{tab:perbench}) shows how these results vary across domains. The largest gains over the IRT~2PL router occur on MuSR ($+12.5$ percentage points) and IFEval ($+14.0$ percentage points).

\begin{wrapfigure}{r}{0.49\textwidth}
\vspace{0mm}
\centering
\includegraphics[width=\linewidth]{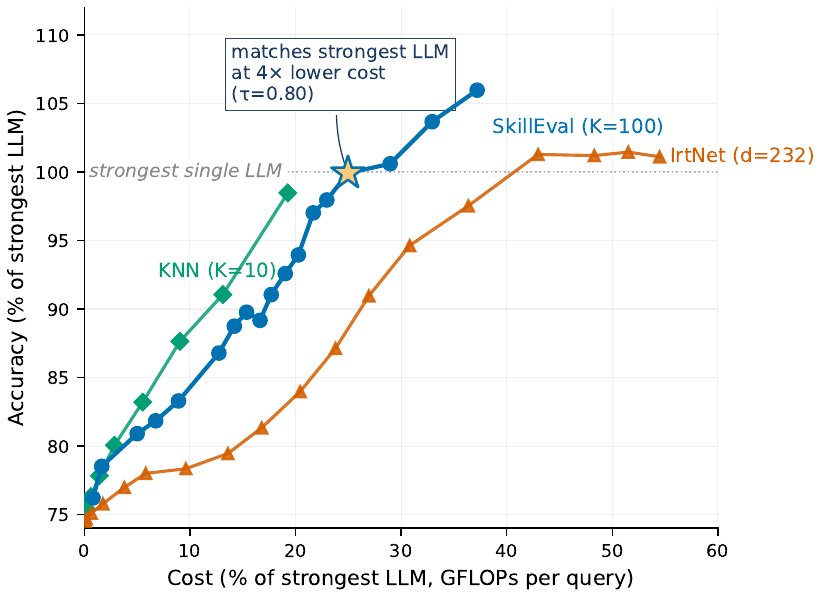}
\caption{Accuracy vs. inference cost for three routers using the same threshold rule. Both axes are expressed relative to the strongest single LLM selected on training items. Each point corresponds to one threshold $\tau$.}
\label{fig:pareto}
\vspace{-3mm}
\end{wrapfigure}
\noindent\textbf{Cost--accuracy trade-off.}\hspace{1em}The preceding experiment selects models for accuracy alone. We now consider inference cost, estimated as the computation per query in billions of floating-point operations (GFLOPs). For each item, the router selects the cheapest LLM whose predicted probability of a correct answer exceeds a threshold $\tau$. If none qualifies, it selects the LLM with the highest predicted probability. Figure~\ref{fig:pareto} compares \textsc{SkillEval}, IRTNet ($d{=}232$), and KNN using this same rule, varying $\tau$ from $0.10$ to $0.95$ and expressing cost relative to the strongest single LLM. At $\tau{=}0.80$, \textsc{SkillEval} reaches $99.9\%$ of that LLM's accuracy at $25\%$ of its cost; at $\tau{=}0.95$, it exceeds that LLM's accuracy by $3.7$ percentage points at $37\%$ of its cost. KNN is more accurate at low cost but levels off at $98\%$ of the strongest LLM's accuracy. Its probability estimates take only the eleven values $0/10,\dots,10/10$, giving many tied candidates at the highest score. At matched cost budgets in the plotted range, \textsc{SkillEval} achieves higher accuracy than IRTNet.

\par\vspace{1.5ex plus 0.5ex minus .2ex}
\noindent\textbf{Skill-level ``weak-beats-strong'' inversions.}\hspace{1em}A fixed ranking of LLMs cannot adapt to questions on which a model with lower average accuracy performs better. To assess how often such opportunities occur, we examine the observed test responses. On $723$ of the $1{,}905$ test items ($38.0\%$), an LLM with at most $13$B parameters answers correctly while the strongest single LLM selected on the training items fails. We call these cases ``weak-beats-strong'' inversions. They occur in every benchmark (BBH: $329$ items, MATH: $156$, GPQA: $155$, MuSR: $64$, IFEval: $19$) and span $41$ skill clusters (Figure~\ref{fig:wbs} in Appendix~\ref{app:wbs}). These cases illustrate why selecting an LLM for each query can be useful even when one model leads on average.

\subsection{Profiling Unseen LLMs}
\label{sec:exp:coldstart}
When a new LLM becomes available, we want to estimate its skill profile from a small set of its answers without retraining the entire diagnostic network. We train the text-conditioned NCDM using only training-item responses from $3{,}048$ LLMs.
We then freeze the network, including the item difficulty and discrimination parameters, and estimate only the $K$-dimensional profile of each new LLM from its answers to $N$ items drawn from the training-item pool. This is ability estimation with fixed item parameters, also called \emph{scoring} in psychometrics \citep{segall1996}.
We evaluate this procedure on $200$ held-out LLMs, using separate test items whose responses never enter profile estimation.

\paragraph{Starting point and related skills.}
We fit a multivariate Gaussian to the raw (pre-sigmoid) profiles of the $3{,}048$ training LLMs and use it as a prior. Starting from the prior mean, we update the new LLM's profile with L-BFGS, using Equation~\ref{eq:ncdm} with a Gaussian prior regularizer. 
Because the Gaussian captures correlations between skills, an answer also shifts skills that are correlated with the item's assigned skills. As more answers are observed, the data term increasingly outweighs the fixed prior (Appendix~\ref{app:coldstart}).

\begin{figure}[t]
\centering
\includegraphics[width=\linewidth]{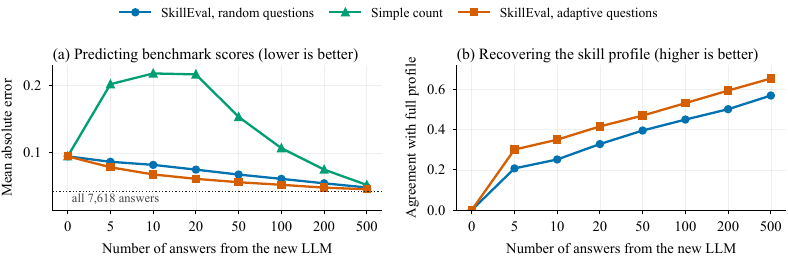}
\caption{Profiling $200$ held-out LLMs from $N$ answers. (a)~Mean absolute benchmark-score error across LLMs and the five benchmarks; the dotted line uses all $7{,}618$ pool answers with the same estimator. (b)~Pearson correlation with profiles estimated from all answers without the prior, computed across LLMs for each skill and averaged over skills.}
\label{fig:profiling}
\end{figure}

\paragraph{Choosing the questions.}
We compare uniform random selection with adaptive selection. Computerized adaptive testing (CAT) chooses each next question based on the answers given so far; CD-CAT applies this idea to cognitive diagnosis models \citep{cheng2009}. Starting from the prior mean, we use the current profile to predict the probability $p_j$ of a correct response to each unused item $j$ and select:
\begin{equation}
j^{*} \;=\; \arg\max_{j}\; p_j\,(1-p_j)\,\|\mathbf{q}_j\|_1 .
\label{eq:cat_heuristic}
\end{equation}
The uncertainty term $p_j(1-p_j)$ is largest at $p_j{=}0.5$.
The skill count $\|\mathbf{q}_j\|_1$ favours items assigned to several skills. After observing the answer, we update the profile using all answers so far and repeat until $N$ items have been answered. Both selection methods use the same final profile-estimation procedure.

\paragraph{Measures.}
We measure what the answers reveal about each LLM in two ways. For each benchmark, we average the predicted probabilities over its test items and compare this predicted accuracy with the observed accuracy on those items. The \emph{benchmark-score error} is their absolute difference, averaged over the five benchmarks and the LLMs. \emph{Profile agreement} compares the relative skill estimates from $N$ answers with a reference profile estimated from all $7{,}618$ pool answers without the prior. We compute Pearson correlation across LLMs for each skill, then average over skills. A \emph{simple-count} baseline uses the same randomly selected items to estimate each benchmark score as the fraction answered correctly. If no items from a benchmark have been answered, it uses the prediction from the zero-answer profile.

\paragraph{Results.}
With no answers, the starting profile predicts benchmark scores with a mean error of $0.095$ (Figure~\ref{fig:profiling}). With $100$ random answers, \textsc{SkillEval} lowers the error to $0.061$, and with $500$ answers to $0.049$, approaching the $0.042$ obtained from all $7{,}618$ answers. Adaptive selection reaches similar errors with fewer answers: its mean error after $10$ answers is $0.068$, comparable to $50$ random answers, and after $20$ answers it is $0.061$, comparable to $100$ random answers. It also yields higher profile agreement: $0.53$ after $100$ answers, versus $0.45$ for random selection. Adaptive selection yields lower benchmark-score error and higher profile agreement than random selection at every tested budget from $5$ to $500$ answers (Appendix~\ref{app:coldstart}). At the nonzero tested budgets up to $100$ answers, the simple count has higher error than the zero-answer prediction; with $500$ answers, its error is $0.052$. 



\section{Diagnostic Analyses}
\label{sec:exp:diagnostic}

The Q-matrix and the learned mastery profiles support complementary analyses of benchmark coverage and model behavior. We examine skill coverage across benchmarks and differences between base and instruction-tuned models. 

\begin{figure}[t]
\centering
\includegraphics[width=\linewidth]{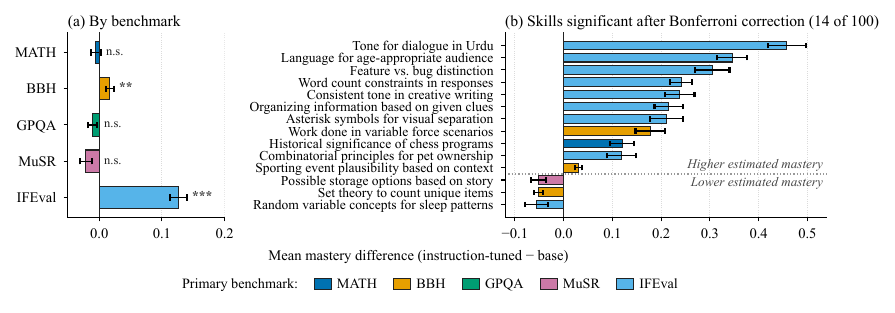}
\caption{Estimated mastery differences between base and instruction-tuned models.
(\textbf{a})~Benchmark means, assigning each skill to its primary benchmark by item count; stars indicate unadjusted $p$-values
($*\,p{<}0.05$, $**\,p{<}0.01$, $***\,p{<}0.001$).
(\textbf{b})~The $14/100$ skills significant after Bonferroni correction
($\alpha{=}5{\times}10^{-4}$), sorted by mean difference and coloured by primary benchmark; names are abbreviated.
Error bars: $\pm1$ SEM across $81$ matched model pairs.
Omitted skills are not necessarily unchanged.}
\label{fig:aligntax}
\vspace{-5mm}
\end{figure}

\paragraph{Skill-level differences between base and instruction-tuned models.}
We compare $81$ matched pairs, each comprising a base model and its instruction-tuned version. Averaged across pairs, estimated mastery is more than $0.01$ higher in instruction-tuned models on $38$ skills and more than $0.01$ lower on $36$ skills. Separately, paired Wilcoxon signed-rank tests identify $14$ skills with significant differences after Bonferroni correction, which divides the significance level by the $100$ tested skills (Figure~\ref{fig:aligntax}). Eleven of these differences are positive, mostly on instruction-following skills, and three are negative, with magnitudes of at most $0.06$. When skills are grouped by the benchmark contributing most of their items, IFEval has the largest mean difference ($\Delta{=}{+}0.126$, $p{<}10^{-10}$); Appendix~\ref{app:alignment} reports the differences by model family. Statistical significance does not establish a large effect, and these observational comparisons do not isolate the causal effect of instruction tuning.

\paragraph{Skill coverage across benchmarks.}
To identify distinct and overlapping skill coverage across benchmarks, we count a skill as covered when at least one item in a benchmark is assigned to it in the Q-matrix. Under this criterion, $30$ of the $100$ skills occur in only one benchmark, including $13$ unique to IFEval, $9$ to GPQA and $6$ to MATH (Figure~\ref{fig:upset} in Appendix~\ref{app:redundancy}). MATH and IFEval together cover $87$ skills; adding GPQA increases this to $98$. BBH adds one further skill to this combined coverage despite contributing $5{,}761$ items. Shared coverage suggests potential redundancy, but benchmarks can still differ in difficulty, item diversity and how well they distinguish between models. The concentration of unique skills in IFEval is consistent with the transfer result in Appendix~\ref{app:transfer}: withholding IFEval from training removes all training items for these skills, and routing performance on IFEval deteriorates.

\section{Conclusion}
\label{sec:conclusion}

\textsc{SkillEval} applies a Neural Cognitive Diagnostic Model to
$3{,}811$ LLMs across $9{,}523$ items and five benchmarks, producing
interpretable mastery profiles over $K{=}100$ named skills. Unlike IRTNet's latent dimensions, these profiles are stable across retraining: each skill's estimates correlate at $0.96$ on average between runs.
Skill-conditioned routing reaches $\mathrm{Acc@}1{=}65.67\%$ (vs.\ $61.57\%$ for the strongest single LLM, $62.05\%$ for KNN, $63.08\%$ for IRTNet over three runs and $61.10\%$ for the IRT~2PL router) and matches the strongest LLM at $4{\times}$ lower inference cost. \textsc{SkillEval} also profiles previously unseen LLMs without retraining the shared network: $20$ adaptively chosen answers give a mean benchmark-score error comparable to $100$ randomly chosen answers. Skill profiles recover
benchmark-level scores at Pearson $r{=}0.989$. The framework surfaces
diagnostic findings invisible to aggregate scores: differences between base and instruction-tuned models that are statistically significant after Bonferroni correction on $14/100$ skills, and overlapping skill coverage across benchmarks. An exploratory analysis in Appendix~\ref{app:comastery_details} examines asymmetric co-mastery: LLMs mastering one skill often also master another, while the reverse is less common. Testing how these patterns generalise across LLM populations is a direction for future work. Limitations are in Appendix~\ref{app:limitations}.

\bibliographystyle{iclr2027_conference}
\bibliography{references}

@article{delatorre2009dina,
author = {Jimmy de la Torre},
title ={DINA Model and Parameter Estimation: A Didactic},

journal = {Journal of Educational and Behavioral Statistics},
volume = {34},
number = {1},
pages = {115-130},
year = {2009},
doi = {10.3102/1076998607309474},

URL = { 
    
        https://doi.org/10.3102/1076998607309474
    
    

},
eprint = { 
    
        https://doi.org/10.3102/1076998607309474
    
    

}
}

@inproceedings{wang2020neural,
    title={Neural Cognitive Diagnosis for Intelligent Education Systems},
    author={Wang, Fei and Liu, Qi and Chen, Enhong and Huang, Zhenya and Chen, Yuying and Yin, Yu and Huang, Zai and Wang, Shijin},
    booktitle={Proceedings of the AAAI Conference on Artificial Intelligence},
    volume={34},
    number={04},
    pages={6153--6161},
    year={2020},
    month={April},
    publisher={Association for the Advancement of Artificial Intelligence (AAAI)},
    doi={10.1609/aaai.v34i04.6080},
    url={https://doi.org/10.1609/aaai.v34i04.6080}
}

@article{wang2022neuralcd,
    title={NeuralCD: A General Framework for Cognitive Diagnosis},
    author={Wang, Fei and Liu, Qi and Chen, Enhong and Huang, Zhenya and Yin, Yu and Wang, Shijin and Su, Yu},
    journal={IEEE Transactions on Knowledge and Data Engineering},
    volume={35},
    number={8},
    pages={8312--8327},
    year={2023},
    month={August},
    publisher={IEEE},
    doi={10.1109/TKDE.2022.3201037}
}

@misc{bigdata2021educdm,
  title={EduCDM},
  author={bigdata-ustc},
  publisher = {GitHub},
  journal = {GitHub repository},
  year = {2021},
  howpublished = {\url{https://github.com/bigdata-ustc/EduCDM}},
}

@misc{gao2023framework,
  author       = {Gao, Leo and Tow, Jonathan and Abbasi, Baber and Biderman, Stella and Black, Sid and DiPofi, Anthony and Foster, Charles and Golding, Laurence and Hsu, Jeffrey and Le Noac'h, Alain and Li, Haonan and McDonell, Kyle and Muennighoff, Niklas and Ociepa, Chris and Phang, Jason and Reynolds, Laria and Schoelkopf, Hailey and Skowron, Aviya and Sutawika, Lintang and Tang, Eric and Thite, Anish and Wang, Ben and Wang, Kevin and Zou, Andy},
  title        = {The Language Model Evaluation Harness},
  month        = 07,
  year         = 2024,
  publisher    = {Zenodo},
  version      = {v0.4.3},
  doi          = {10.5281/zenodo.12608602},
  url          = {https://zenodo.org/records/12608602}
}

@misc{hu2024routerbench,
      title={RouterBench: A Benchmark for Multi-LLM Routing System}, 
      author={Qitian Jason Hu and Jacob Bieker and Xiuyu Li and Nan Jiang and Benjamin Keigwin and Gaurav Ranganath and Kurt Keutzer and Shriyash Kaustubh Upadhyay},
      year={2024},
      eprint={2403.12031},
      archivePrefix={arXiv},
      primaryClass={cs.LG},
      url={https://arxiv.org/abs/2403.12031}, 
}

@misc{didolkar2024metacognitive,
      title={Metacognitive Capabilities of LLMs: An Exploration in Mathematical Problem Solving}, 
      author={Aniket Didolkar and Anirudh Goyal and Nan Rosemary Ke and Siyuan Guo and Michal Valko and Timothy Lillicrap and Danilo Rezende and Yoshua Bengio and Michael Mozer and Sanjeev Arora},
      year={2024},
      eprint={2405.12205},
      archivePrefix={arXiv},
      primaryClass={cs.AI},
      url={https://arxiv.org/abs/2405.12205}
}

@inproceedings{song2025irtrouter,
    title = "{IRT}-Router: Effective and Interpretable Multi-{LLM} Routing via Item Response Theory",
    author = "Song, Wei  and
      Huang, Zhenya  and
      Cheng, Cheng  and
      Gao, Weibo  and
      Xu, Bihan  and
      Zhao, GuanHao  and
      Wang, Fei  and
      Wu, Runze",
    editor = "Che, Wanxiang  and
      Nabende, Joyce  and
      Shutova, Ekaterina  and
      Pilehvar, Mohammad Taher",
    booktitle = "Proceedings of the 63rd Annual Meeting of the Association for Computational Linguistics (Volume 1: Long Papers)",
    month = jul,
    year = "2025",
    address = "Vienna, Austria",
    publisher = "Association for Computational Linguistics",
    url = "https://aclanthology.org/2025.acl-long.761/",
    doi = "10.18653/v1/2025.acl-long.761",
    pages = "15629--15644",
    ISBN = "979-8-89176-251-0"
}

@misc{chen2025irtnet,
      title={Learning Compact Representations of LLM Abilities via Item Response Theory}, 
      author={Jianhao Chen and Chenxu Wang and Gengrui Zhang and Peng Ye and Lei Bai and Wei Hu and Yuzhong Qu and Shuyue Hu},
      year={2025},
      eprint={2510.00844},
      archivePrefix={arXiv},
      primaryClass={cs.AI},
      url={https://arxiv.org/abs/2510.00844}, 
}

@inproceedings{hendrycks2021measuring,
 author = {Hendrycks, Dan and Burns, Collin and Kadavath, Saurav and Arora, Akul and Basart, Steven and Tang, Eric and Song, Dawn and Steinhardt, Jacob},
 booktitle = {Proceedings of the Neural Information Processing Systems Track on Datasets and Benchmarks},
 editor = {J. Vanschoren and S. Yeung},
 pages = {},
 title = {Measuring Mathematical Problem Solving With the MATH Dataset},
 url = {https://datasets-benchmarks-proceedings.neurips.cc/paper_files/paper/2021/file/be83ab3ecd0db773eb2dc1b0a17836a1-Paper-round2.pdf},
 volume = {1},
 year = {2021}
}

@misc{cobbe2021training,
      title={Training Verifiers to Solve Math Word Problems}, 
      author={Karl Cobbe and Vineet Kosaraju and Mohammad Bavarian and Mark Chen and Heewoo Jun and Lukasz Kaiser and Matthias Plappert and Jerry Tworek and Jacob Hilton and Reiichiro Nakano and Christopher Hesse and John Schulman},
      year={2021},
      eprint={2110.14168},
      archivePrefix={arXiv},
      primaryClass={cs.LG},
      url={https://arxiv.org/abs/2110.14168}, 
}

@inproceedings{suzgun-etal-2023-challenging,
    title = "Challenging {BIG}-Bench Tasks and Whether Chain-of-Thought Can Solve Them",
    author = {Suzgun, Mirac  and
      Scales, Nathan  and
      Sch{\"a}rli, Nathanael  and
      Gehrmann, Sebastian  and
      Tay, Yi  and
      Chung, Hyung Won  and
      Chowdhery, Aakanksha  and
      Le, Quoc  and
      Chi, Ed  and
      Zhou, Denny  and
      Wei, Jason},
    editor = "Rogers, Anna  and
      Boyd-Graber, Jordan  and
      Okazaki, Naoaki",
    booktitle = "Findings of the Association for Computational Linguistics: ACL 2023",
    month = jul,
    year = "2023",
    address = "Toronto, Canada",
    publisher = "Association for Computational Linguistics",
    url = "https://aclanthology.org/2023.findings-acl.824/",
    doi = "10.18653/v1/2023.findings-acl.824",
    pages = "13003--13051"
}

@inproceedings{cui-sachan-2025-investigating,
    title = "Investigating the Zone of Proximal Development of Language Models for In-Context Learning",
    author = "Cui, Peng  and
      Sachan, Mrinmaya",
    editor = "Chiruzzo, Luis  and
      Ritter, Alan  and
      Wang, Lu",
    booktitle = "Findings of the Association for Computational Linguistics: NAACL 2025",
    month = apr,
    year = "2025",
    address = "Albuquerque, New Mexico",
    publisher = "Association for Computational Linguistics",
    url = "https://aclanthology.org/2025.findings-naacl.362/",
    doi = "10.18653/v1/2025.findings-naacl.362",
    pages = "6485--6498",
    ISBN = "979-8-89176-195-7"
}

@inproceedings{gor-etal-2024-great,
    title = "Do great minds think alike? Investigating Human-{AI} Complementarity in Question Answering with {CAIMIRA}",
    author = "Gor, Maharshi  and
      Daum{\'e} III, Hal  and
      Zhou, Tianyi  and
      Boyd-Graber, Jordan",
    editor = "Al-Onaizan, Yaser  and
      Bansal, Mohit  and
      Chen, Yun-Nung",
    booktitle = "Proceedings of the 2024 Conference on Empirical Methods in Natural Language Processing",
    month = nov,
    year = "2024",
    address = "Miami, Florida, USA",
    publisher = "Association for Computational Linguistics",
    url = "https://aclanthology.org/2024.emnlp-main.1201/",
    doi = "10.18653/v1/2024.emnlp-main.1201",
    pages = "21533--21564"
}

@inproceedings{
kipnis2025metabench,
title={metabench - A Sparse Benchmark of Reasoning and Knowledge in Large Language Models},
author={Alex Kipnis and Konstantinos Voudouris and Luca M. Schulze Buschoff and Eric Schulz},
booktitle={The Thirteenth International Conference on Learning Representations},
year={2025},
url={https://openreview.net/forum?id=4T33izzFpK}
}

@article{zouhar-etal-2025-select,
    title = "How to Select Datapoints for Efficient Human Evaluation of {NLG} Models?",
    author = "Zouhar, Vil{\'e}m  and
      Cui, Peng  and
      Sachan, Mrinmaya",
    journal = "Transactions of the Association for Computational Linguistics",
    volume = "13",
    year = "2025",
    address = "Cambridge, MA",
    publisher = "MIT Press",
    url = "https://aclanthology.org/2025.tacl-1.80/",
    doi = "10.1162/tacl.a.60",
    pages = "1789--1811",
}

@inproceedings{rodriguez-etal-2021-evaluation,
    title = "Evaluation Examples are not Equally Informative: How should that change {NLP} Leaderboards?",
    author = "Rodriguez, Pedro  and
      Barrow, Joe  and
      Hoyle, Alexander  and
      Lalor, John P.  and
      Jia, Robin  and
      Boyd-Graber, Jordan",
    editor = "Zong, Chengqing  and
      Xia, Fei  and
      Li, Wenjie  and
      Navigli, Roberto",
    booktitle = "Proceedings of the 59th Annual Meeting of the Association for Computational Linguistics and the 11th International Joint Conference on Natural Language Processing (Volume 1: Long Papers)",
    month = aug,
    year = "2021",
    address = "Online",
    publisher = "Association for Computational Linguistics",
    url = "https://aclanthology.org/2021.acl-long.346/",
    doi = "10.18653/v1/2021.acl-long.346",
    pages = "4486--4503",
}

@incollection{reckase200618,
title = {18 Multidimensional Item Response Theory},
editor = {C.R. Rao and S. Sinharay},
series = {Handbook of Statistics},
publisher = {Elsevier},
volume = {26},
pages = {607-642},
year = {2006},
booktitle = {Psychometrics},
issn = {0169-7161},
doi = {https://doi.org/10.1016/S0169-7161(06)26018-8},
url = {https://www.sciencedirect.com/science/article/pii/S0169716106260188},
author = {Mark D. Reckase},
}

@book{lord2008statistical,
    title={Statistical Theories of Mental Test Scores},
    author={Lord, Frederic M. and Novick, Melvin R.},
    year={1968},
    publisher={Addison-Wesley},
}

@inproceedings{
rein2024gpqa,
title={{GPQA}: A Graduate-Level Google-Proof Q\&A Benchmark},
author={David Rein and Betty Li Hou and Asa Cooper Stickland and Jackson Petty and Richard Yuanzhe Pang and Julien Dirani and Julian Michael and Samuel R. Bowman},
booktitle={First Conference on Language Modeling},
year={2024},
url={https://openreview.net/forum?id=Ti67584b98}
}

@inproceedings{
sprague2024musr,
title={Mu{SR}: Testing the Limits of Chain-of-thought with Multistep Soft Reasoning},
author={Zayne Rea Sprague and Xi Ye and Kaj Bostrom and Swarat Chaudhuri and Greg Durrett},
booktitle={The Twelfth International Conference on Learning Representations},
year={2024},
url={https://openreview.net/forum?id=jenyYQzue1}
}

@misc{zhou2023instruction,
      title={Instruction-Following Evaluation for Large Language Models}, 
      author={Jeffrey Zhou and Tianjian Lu and Swaroop Mishra and Siddhartha Brahma and Sujoy Basu and Yi Luan and Denny Zhou and Le Hou},
      year={2023},
      eprint={2311.07911},
      archivePrefix={arXiv},
      primaryClass={cs.CL},
      url={https://arxiv.org/abs/2311.07911}, 
}

@inproceedings{huang2025routereval,
    title = "{R}outer{E}val: A Comprehensive Benchmark for Routing {LLM}s to Explore Model-level Scaling Up in {LLM}s",
    author = "Huang, Zhongzhan  and
      Ling, Guoming  and
      Lin, Yupei  and
      Chen, Yandong  and
      Zhong, Shanshan  and
      Wu, Hefeng  and
      Lin, Liang",
    editor = "Christodoulopoulos, Christos  and
      Chakraborty, Tanmoy  and
      Rose, Carolyn  and
      Peng, Violet",
    booktitle = "Findings of the Association for Computational Linguistics: EMNLP 2025",
    month = nov,
    year = "2025",
    address = "Suzhou, China",
    publisher = "Association for Computational Linguistics",
    url = "https://aclanthology.org/2025.findings-emnlp.208/",
    doi = "10.18653/v1/2025.findings-emnlp.208",
    pages = "3860--3887",
}

@misc{biderman2024lessons,
      title={Lessons from the Trenches on Reproducible Evaluation of Language Models}, 
      author={Stella Biderman and Hailey Schoelkopf and Lintang Sutawika and Leo Gao and Jonathan Tow and Baber Abbasi and Alham Fikri Aji and Pawan Sasanka Ammanamanchi and Sidney Black and Jordan Clive and Anthony DiPofi and Julen Etxaniz and Benjamin Fattori and Jessica Zosa Forde and Charles Foster and Jeffrey Hsu and Mimansa Jaiswal and Wilson Y. Lee and Haonan Li and Charles Lovering and Niklas Muennighoff and Ellie Pavlick and Jason Phang and Aviya Skowron and Samson Tan and Xiangru Tang and Kevin A. Wang and Genta Indra Winata and François Yvon and Andy Zou},
      year={2024},
      eprint={2405.14782},
      archivePrefix={arXiv},
      primaryClass={cs.CL},
      url={https://arxiv.org/abs/2405.14782}, 
}

@inproceedings{polo2024tinybenchmarks,
  title = 	 {tiny{B}enchmarks: evaluating {LLM}s with fewer examples},
  author =       {Maia Polo, Felipe and Weber, Lucas and Choshen, Leshem and Sun, Yuekai and Xu, Gongjun and Yurochkin, Mikhail},
  booktitle = 	 {Proceedings of the 41st International Conference on Machine Learning},
  pages = 	 {34303--34326},
  year = 	 {2024},
  editor = 	 {Salakhutdinov, Ruslan and Kolter, Zico and Heller, Katherine and Weller, Adrian and Oliver, Nuria and Scarlett, Jonathan and Berkenkamp, Felix},
  volume = 	 {235},
  series = 	 {Proceedings of Machine Learning Research},
  month = 	 {21--27 Jul},
  publisher =    {PMLR},
  url = 	 {https://proceedings.mlr.press/v235/maia-polo24a.html}
}

@article{schonemann1966,
  author = {Sch{\"o}nemann, Peter H.},
  title = {A generalized solution of the orthogonal procrustes problem},
  journal = {Psychometrika},
  year = {1966},
  volume = {31},
  number = {1},
  pages = {1--10},
  doi = {10.1007/BF02289451},
  url = {https://doi.org/10.1007/BF02289451}
}

@article{gower1975,
  author = {Gower, J. C.},
  title = {Generalized procrustes analysis},
  journal = {Psychometrika},
  year = {1975},
  volume = {40},
  number = {1},
  pages = {33--51},
  doi = {10.1007/BF02291478},
  url = {https://doi.org/10.1007/BF02291478}
}

@Article{cheng2009,
  author = {Cheng, Ying},
  title = {When Cognitive Diagnosis Meets Computerized Adaptive Testing: CD-CAT},
  journal = {Psychometrika},
  year = {2009},
  month = {Dec},
  day = {01},
  volume = {74},
  number = {4},
  pages = {619--632},
  issn = {1860-0980},
  doi = {10.1007/s11336-009-9123-2},
  url = {https://doi.org/10.1007/s11336-009-9123-2}
}

@Article{segall1996,
  author = {Segall, Daniel O.},
  title = {Multidimensional adaptive testing},
  journal = {Psychometrika},
  year = {1996},
  month = {Jun},
  day = {01},
  volume = {61},
  number = {2},
  pages = {331--354},
  issn = {1860-0980},
  doi = {10.1007/BF02294343},
  url = {https://doi.org/10.1007/BF02294343}
}

@inproceedings{
he2026stat,
title={{STAT}: Skill-Targeted Adaptive Training},
author={Yinghui He and Abhishek Panigrahi and Yong Lin and Sanjeev Arora},
booktitle={The Fourteenth International Conference on Learning Representations},
year={2026},
url={https://openreview.net/forum?id=m3jG3GaNIj}
}

@inproceedings{reimers-gurevych-2019-sentence,
    title = "Sentence-{BERT}: Sentence Embeddings using {S}iamese {BERT}-Networks",
    author = "Reimers, Nils  and
      Gurevych, Iryna",
    editor = "Inui, Kentaro  and
      Jiang, Jing  and
      Ng, Vincent  and
      Wan, Xiaojun",
    booktitle = "Proceedings of the 2019 Conference on Empirical Methods in Natural Language Processing and the 9th International Joint Conference on Natural Language Processing (EMNLP-IJCNLP)",
    month = nov,
    year = "2019",
    address = "Hong Kong, China",
    publisher = "Association for Computational Linguistics",
    url = "https://aclanthology.org/D19-1410/",
    doi = "10.18653/v1/D19-1410",
    pages = "3982--3992"
}

@misc{cui2026makesharderexplainingquestion,
      title={What Makes Something Hard(er)? Explaining Question Difficulty in Natural Language}, 
      author={Peng Cui and Qiaoyuan Zheng and Rudolf Debelak and Mrinmaya Sachan},
      year={2026},
      eprint={2610.01627},
      archivePrefix={arXiv},
      primaryClass={cs.CL},
      url={https://arxiv.org/abs/2610.01627}, 
}

@misc{cui2026llmsexhibitcoherentknowledge,
      title={Do LLMs Exhibit Coherent Knowledge Structures in Mathematical Reasoning? A Perspective from Knowledge Space Theory}, 
      author={Peng Cui and Heejin Do and Mrinmaya Sachan},
      year={2026},
      eprint={2609.05245},
      archivePrefix={arXiv},
      primaryClass={cs.AI},
      url={https://arxiv.org/abs/2609.05245}, 
}

\appendix

\section{Related Work}
\label{app:related}

\paragraph{LLM Evaluation and Routing.}
Standard LLM evaluation aggregates item-level correctness into a single
benchmark score \citep{gao2023framework, hendrycks2021measuring,
cobbe2021training}, creating leaderboards whose rankings are influenced as
much by evaluation protocol (prompt format, shot count, decoding parameters)
as by true model capability~\citep{biderman2024lessons}. Platforms such as the
Open LLM Leaderboard and RouterBench \cite{hu2024routerbench} release
per-item correctness matrices for thousands of LLMs, enabling but not
themselves performing fine-grained analysis. Routing systems match queries
to the cheapest capable model, typically using latent query-model embeddings
without explicit semantic skills. Closest in spirit to our work, CAIMIRA
\cite{gor-etal-2024-great} applies IRT to factual QA, and \cite{didolkar2024metacognitive} use LLM-generated skill labels to probe
mathematical reasoning. We build on the latter by scaling skill extraction
to five benchmarks and fitting a full CDM over the extracted skills.

\paragraph{Cognitive Diagnostic Models.}
Cognitive Diagnostic Models (CDMs) originate in educational psychometrics
and estimate latent skill mastery from observed item responses, extending
Item Response Theory (IRT) by modelling proficiency as a vector over
\emph{named} skills rather than a scalar or a continuous embedding.
Classical CDMs such as DINA \cite{delatorre2009dina} assume conjunctive
skill requirements (an item is correct only if all required skills are
mastered) and binary mastery. The Neural CDM (NCDM) \cite{wang2020neural}
relaxes these assumptions with neural interaction functions while
preserving mastery-to-score monotonicity via non-negative weight
constraints, and is generalised in NeuralCD \cite{wang2022neuralcd};
reference implementations are collected in the EduCDM library
\cite{bigdata2021educdm}. CDMs have been widely deployed for human
assessment over hundreds of skills and tens of thousands of examinees,
making them a natural fit for LLM evaluation where response matrices are
now orders of magnitude larger.

\paragraph{Psychometric Methods for LLMs.}
Recent work applies psychometric models to LLMs at scale. IRT-Router
\cite{song2025irtrouter} uses multidimensional IRT with a query-cluster
relevance vector for routing. IRTNet \cite{chen2025irtnet} jointly learns
$d$-dimensional LLM ability embeddings, query discrimination, and query
difficulty via a Mixture-of-Experts architecture, evaluated on $112$ LLMs
and $10$ benchmarks. Metabench~\citep{kipnis2025metabench} compresses six public
benchmarks to under $3\%$ of their combined items using $2$-PL IRT and
Fisher-information filtering on ${\sim}\,5{,}000$ LLMs, showing that a
single latent factor explains $79\%$ of benchmark variance. TinyBenchmarks~\citep{polo2024tinybenchmarks}
similarly uses IRT-based item selection to reduce evaluation cost. All of these approaches treat LLM ability as a
low-dimensional latent embedding without a named skill structure.
\textsc{SkillEval} commits to an \emph{a priori} Q-matrix over $K{=}100$
semantically labeled skills derived from LLM metacognition and SBERT
clustering, enabling skill-level analyses of co-mastery patterns,
skill interaction within items, and base versus instruction-tuned differences that
are not expressible over a continuous latent embedding. We scale to
$3{,}811$ LLMs ($34{\times}$ IRTNet, comparable to metabench's
${\sim}\,5{,}000$) across five benchmarks.

\section{Notation}
\label{app:notation}

Table~\ref{tab:notation} summarises the notation used throughout the paper.

\begin{table}[h]
\centering
\small
\caption{Notation summary.}
\label{tab:notation}
\begin{tabular}{ll}
\toprule
\textbf{Symbol} & \textbf{Meaning} \\
\midrule
$M$ & Number of LLMs ($M{=}3{,}811$) \\
$N$ & Number of items ($N{=}9{,}523$); also the number of answers used to profile a new LLM \\
$K$ & Number of skill clusters ($K{=}100$ default) \\
$i\in\{1,\dots,M\}$ & LLM index \\
$j\in\{1,\dots,N\}$ & Item index \\
$k\in\{1,\dots,K\}$ & Skill index \\
$\mathcal{S}\in\{0,1\}^{M\times N}$ & Binary response matrix; $\mathcal{S}_{ij}{=}1$ iff LLM $i$ answers item $j$ correctly \\
$\mathcal{Q}\in\{0,1\}^{N\times K}$ & Q-matrix; $q_{jk}{=}1$ iff item $j$ tests skill $k$ \\
$\boldsymbol{\theta}_i\in[0,1]^K$ & Skill mastery profile of LLM $i$ \\
$\boldsymbol{d}_j\in[0,1]^K$ & Item difficulty per skill \\
$\alpha_j\in[0,1]$ & Item discrimination \\
$\boldsymbol{e}_j\in\R^{768}$ & Frozen SBERT text embedding of item $j$ \\
$\sigma(\cdot)$ & Sigmoid function \\
\bottomrule
\end{tabular}
\end{table}

\section{Skill Cluster Examples}
\label{app:clusters}

Table~\ref{tab:clusters} shows four representative skill clusters with example
items. Each cluster groups items that share a common cognitive requirement
despite coming from different benchmarks or subtasks. For instance, the
\emph{Quadratic Formula} cluster contains $451$ items dominated by MATH
algebra problems but also includes $20$ GPQA items that happen to require
algebraic manipulation.

\begin{table*}[t]
\centering
\small
\caption{Representative skill clusters with example items. Each cluster
aggregates items from multiple benchmarks that share a common skill
requirement.}
\label{tab:clusters}
\begin{tabular}{p{3.2cm}rp{8.7cm}}
\toprule
\textbf{Skill cluster} & \textbf{Items} & \textbf{Example items} \\
\midrule
Quadratic Formula (MATH-primary, 429/451) & 451 &
  ``Find the domain of $f(x)=\sqrt{\sqrt{x^2-16}-3}$''
  \newline ``If $-6\leq a\leq -2$ and $3\leq b\leq 5$, find the greatest value of $\frac{a+b}{a}$'' \\
\midrule
Truthfulness of Nested Statements (BBH-primary, 1836/1880) & 1{,}880 &
  ``Then he tells Billy: `Be sure that you do not come in at all tomorrow morning'\,'' (causal judgement)
  \newline ``Teressa lies. Willian says Teressa lies. Jamey says Willian lies. Ryan says Jamey lies. Kandi says Ryan lies. Does Kandi tell the truth?'' (web of lies) \\
\midrule
Structural Isomers in Organic Compounds (GPQA-primary, 442/488) & 488 &
  ``Identify product D [from NMR spectrum]: (A)~2-bromonaphthalene (B)~2-bromo-1,2,3,4-tetrahydronaphthalene\ldots'' \\
\midrule
Word Count Constraints (IFEval-primary, 414/418) & 418 &
  ``Write a social media post about the hajj pilgrimage. Use all lowercase and include the word `story' at least twice'' \\
\bottomrule
\end{tabular}
\end{table*}

\paragraph{Divergent model performance.}
Table~\ref{tab:divergent} illustrates how skill profiles explain item-level
performance differences. On a BBH book-trading problem that requires both
\emph{sequential event tracking} and \emph{object trade tracking}
(conjunctive), models mastering both skills solve it $74\%$ of the time
while models mastering only one skill drop to $22$--$33\%$. For example,
Llama-3.1-Tulu-3-70B (masters both) answers correctly, while
Llama-3.2-Korean-5B (masters only event tracking) fails. On a compensatory
IFEval item requiring \emph{word count constraints} or \emph{Urdu dialogue
adaptation}, mastering either skill suffices: accuracy is $97\%$ with word
count skill alone and $42\%$ with Urdu skill alone, both far above the
$12\%$ baseline.

\begin{table}[t]
\centering
\small
\caption{Accuracy by mastery group on two held-out items. The BBH item is
conjunctive (needing both skills). The IFEval item is compensatory (either
skill suffices).}
\label{tab:divergent}
\begin{tabular}{lcccc}
\toprule
& \textbf{Both} & \textbf{Skill 1} & \textbf{Skill 2} & \textbf{Neither} \\
\midrule
\textbf{BBH} (conjunctive) & 0.74 & 0.33 & 0.22 & 0.10 \\
\textbf{IFEval} (compensatory) & 0.93 & 0.97 & 0.42 & 0.12 \\
\bottomrule
\end{tabular}
\end{table}

\section{Q-Matrix Construction Details}
\label{app:qmatrix}

\paragraph{Skill extraction prompt.}
For each item we prompt GPT-4o-mini at temperature $0$ with the benchmark, the subtask and the item text, and ask for two to four skills, each with a label, the cognitive process and the content domain, together with the single most important skill, which we call the primary skill. Labels on the list of generic labels in the prompt are removed after extraction. The system prompt and the user template are reproduced below; long lines are wrapped for display. We extract the full skill list rather than only the primary skill, yielding a mean of $2.9$ skills per
item and $13{,}276$ unique skill strings across $9{,}523$ items. The primary
skill alone is too coarse: it covers only the headline category and produces
Q-matrix rows with mean row-sum $1.0$, which under-constrains the diagnostic
model.

\noindent\textit{System prompt.}
\begin{footnotesize}
\begin{verbatim}
You are a psychometrician designing a Q-matrix for cognitive diagnostic
   assessment of AI systems. Your task: identify the specific cognitive
   skills each test item measures.

RULES:
1. Each skill must be a compound phrase (5-10 words):
   [specific_cognitive_process] + [specific_content_domain]
2. Extract 2-4 skills per item. At least one must be domain-specific.
3. FORBIDDEN generic labels (will be rejected):
   - "reading comprehension", "logical reasoning", "critical thinking"
   - "mathematical reasoning", "problem solving", "analytical thinking"
   - "scientific knowledge", "attention to detail", "numerical
      reasoning"
   - Any label that would apply to >15% of items in the benchmark
4. Specificity test: would fewer than 100 items in a 10,000-item test
   share this exact skill? If not, decompose further.
5. Use lowercase snake_case labels.

GOOD examples by benchmark type:

MATH: "factoring_higher_degree_polynomials_over_integers",
   "applying_pigeonhole_principle_to_combinatorial_bounds",
   "computing_modular_arithmetic_in_residue_classes"

BBH: "tracing_boolean_operator_precedence_in_nested_expressions",
   "identifying_causal_direction_from_correlational_evidence",
   "tracking_object_positions_through_spatial_transformations"

GPQA: "applying_gauss_law_to_cylindrical_charge_distributions",
   "predicting_reaction_products_via_retrosynthetic_analysis",
   "interpreting_phylogenetic_trees_from_molecular_sequence_data"

MuSR: "integrating_alibis_and_motives_to_identify_suspects",
   "tracking_object_locations_across_sequential_room_transfers",
   "resolving_team_allocation_under_mutual_exclusion_constraints"

IFEval: "enforcing_exact_word_count_in_structured_output",
   "maintaining_consistent_formatting_across_nested_lists",
   "embedding_required_keywords_while_preserving_coherence"

BAD examples (too generic, rejected):
- "problem solving" -> decompose to
   "applying_substitution_to_solve_nonlinear_systems"
- "reading comprehension" -> decompose to
   "extracting_temporal_ordering_from_narrative_clues"
- "scientific knowledge" -> decompose to
   "applying_conservation_of_angular_momentum_to_rotating_bodies"
\end{verbatim}
\end{footnotesize}

\noindent\textit{User template.} The benchmark name, subtask and item text replace the placeholders in braces.
\begin{footnotesize}
\begin{verbatim}
Benchmark: {benchmark} | Subtask: {subtask}

Item:
{question_text}

Respond in JSON:
{
  "skills": [
    {
      "label": "5-10 word compound skill in snake_case",
      "process": "the specific cognitive operation",
      "domain": "the specific content area"
    }
  ],
  "primary_skill": "label of the single most important skill"
}
\end{verbatim}
\end{footnotesize}

\paragraph{Embedding and clustering.}
The $13{,}276$ unique skill strings are embedded with the
\texttt{all-mpnet-base-v2} sentence transformer into $\R^{768}$ and
$L_2$-normalised. We compare three clustering families: HDBSCAN after
UMAP reduction to $5$D, $K$-Means, and HAC with cosine distance and
average linkage. We select HAC because it yields the highest downstream
test AUC at every $K$ we tried, despite a lower silhouette score
($0.053$ at $K{=}50$ vs.\ $0.144$ for $K$-Means and $0.160$ for HDBSCAN
at $K{=}281$). Intrinsic clustering quality is therefore not a reliable
proxy for diagnostic utility.

\paragraph{Coverage fix.}
Our first pipeline iteration assigned each item to a single cluster via the
primary skill only, which left $43\%$ of clusters with fewer than $10$ items and $10$ empty columns in $\mathbf{Q}$ at $K{=}100$. Switching to the multi-skill assignment described above removed all empty columns, reduced the fraction of under-populated clusters to $36\%$ and raised the mean column-sum from $95$ to $139$ items per skill.

\paragraph{Reference-answer ablation.}
Our default skill extraction does not use reference solutions. To test whether showing the extractor a reference solution changes the result, we ran two new extractions for the $2{,}470$ MATH and GPQA items that have a reference solution, one from the question and one from the question plus the solution. Both use the same prompt, the same extraction model and the same question text, and they are separate from the extraction that produced the Q-matrix. The solution changes which skills are named: the token-level Jaccard overlap between the two label sets is $0.37$, against $0.73$ between two independent question-only runs. A blind judge (GPT-4.1-mini) compared both label sets on $600$ items, each in both presentation orders. We retain the $357$ items with a consistent, non-tied preference in both orders; the remaining judgments do not enter the percentages below. Among the retained items the judge preferred the solution-aware labels in $49.6\%$ and the question-only labels in $50.4\%$, so it showed no clear aggregate preference for either extraction condition. Within the retained subset the preference differs by benchmark: question-only labels are preferred on MATH ($59\%$) and solution-aware labels on GPQA ($61\%$). We keep question-only extraction, which also applies to item banks without reference solutions.

\section{Training Details}
\label{app:training}

The text-conditioned NCDM is trained end-to-end with binary cross-entropy loss on the $80\%$ training split of the response matrix, which covers $7{,}618$ training items and all $3{,}811$ LLMs. The skill mastery embedding $\boldsymbol{\theta}_i$, the item difficulty and discrimination heads $\mathbf{W}_k\in\R^{K\times 768}$ and $\mathbf{w}_e\in\R^{768}$, and the PosLinear layers all use Xavier normal initialisation. The PosLinear prediction network has hidden widths $K{\to}512{\to}256{\to}1$; non-negativity is enforced by $W^{+}{=}\max(W,0)$ at each forward pass. We optimise with Adam at learning rate $0.002$ for $15$ epochs, with dropout $0.5$ after each hidden layer. The main model and the $K$-sweep models use batch size $1024$. The five-seed runs and the interaction-wise model use batch size $64$. Model selection uses validation AUC on a held-out triplet split of $10\%$ of the training pairs, keeping the weights of the best-AUC epoch. Training the main model takes about three hours on a single NVIDIA RTX~2080~Ti. The default seed is $42$. The multi-seed results in Section~\ref{sec:stability} use seeds $42$--$46$.

\section{Full K-Sweep Results}
\label{app:ksweep}

Table~\ref{tab:ksweep} reports a sweep over $K\in\{50,100,150,200,300,400,500\}$.
Test AUC is essentially flat (0.716--0.721), confirming that the diagnostic
fit is robust to clustering granularity rather than dependent on a particular
elbow. Routing $\mathrm{Acc@}1$ improves modestly from $0.640$ at $K{=}50$ to
$0.676$ at $K{=}500$, and is $0.657$ at $K{=}100$. We adopt $K{=}100$ as the default because it preserves
interpretable, named clusters while using about four times fewer parameters than
$K{=}500$.

\begin{table}[t]
\centering
\small
\caption{Effect of the number of skill clusters $K$ on diagnostic AUC and
routing Acc@$k$. Diagnostic fit is essentially constant across $K$, and routing
accuracy rises modestly with $K$. Each row is a single training run with batch size $1024$; the $K{=}100$ row is the main model, so its test AUC differs from the five-seed mean in Table~\ref{tab:item_prediction}.}
\label{tab:ksweep}
\begin{tabular}{rccccc}
\toprule
$K$ & \textbf{Test AUC} & \textbf{Acc@1} & \textbf{Acc@3} & \textbf{Acc@5} & \textbf{Acc@10} \\
\midrule
50  & 0.720 & 0.640 & 0.783 & 0.829 & 0.877 \\
100 & 0.717 & 0.657 & 0.780 & 0.829 & 0.875 \\
150 & 0.719 & 0.654 & 0.778 & 0.831 & 0.887 \\
200 & 0.716 & 0.661 & 0.797 & 0.846 & 0.885 \\
300 & 0.719 & 0.672 & 0.794 & 0.843 & 0.882 \\
400 & 0.721 & 0.669 & 0.802 & 0.841 & 0.887 \\
500 & 0.720 & 0.676 & 0.804 & 0.847 & 0.896 \\
\bottomrule
\end{tabular}
\end{table}

\section{Profiling Unseen LLMs: Details}
\label{app:coldstart}

\textbf{Network.} The frozen network is trained with the recipe of the main model in Appendix~\ref{app:training} (seed $42$, batch size $1024$, $15$ epochs), using only the training-item responses of the $3{,}048$ training LLMs.

\textbf{Estimation.} Each profile is represented by unconstrained raw parameters $\mathbf{r}$, with $\boldsymbol{\theta}=\mathrm{sigmoid}(\mathbf{r})$. The prior on $\mathbf{r}$ is a multivariate normal distribution. Its mean $\boldsymbol{\mu}$ is the mean raw profile of the $3{,}048$ training LLMs; its covariance $\boldsymbol{\Sigma}$ is their empirical covariance shrunk by $10\%$ towards its diagonal. Let $A$ be the set of answered items, $y_j$ indicate whether item $j$ was answered correctly, and $p_j$ be the predicted probability of correctness from Equation~\ref{eq:ncdm}. We estimate $\mathbf{r}$ by minimising
\begin{equation}
-\sum_{j\in A}\big[y_j\log p_j+(1-y_j)\log(1-p_j)\big]\;+\;\tfrac{1}{2}(\mathbf{r}-\boldsymbol{\mu})^{\top}\boldsymbol{\Sigma}^{-1}(\mathbf{r}-\boldsymbol{\mu}).
\end{equation}
The first term measures fit to the observed answers; the second penalises departures from profiles supported by the prior. We optimise with L-BFGS, starting from $\boldsymbol{\mu}$. With no answers, the raw estimate is $\boldsymbol{\mu}$ and the skill profile is its sigmoid transform. In a check on the first $20$ LLMs using all $7{,}618$ pool answers, fits with and without the prior have a mean absolute difference of $0.005$ in their test-item probabilities. During adaptive selection, each update starts from the previous raw estimate. At each evaluated budget, we refit from $\boldsymbol{\mu}$, using the same procedure for random and adaptive selection.

\textbf{Evaluation.} We use the first $200$ of the $763$ held-out LLMs. Random-selection metrics are averaged over three seeds; adaptive selection is deterministic for a given LLM and network. Both selection methods are evaluated at the same answer budgets up to $500$. Table~\ref{tab:profiling} lists the values shown in Figure~\ref{fig:profiling}. At each tested budget from $5$ to $500$ answers, paired $95\%$ bootstrap intervals over LLMs favour adaptive selection on both measures: the adaptive-minus-random difference is below zero for benchmark-score error and above zero for profile agreement.

\textbf{Reference profiles and robustness.} Profile agreement uses a reference estimated from all pool answers without the prior, avoiding a shared prior term in the reference and the estimates being evaluated. This reference is itself an estimated profile. We assign zero agreement to a skill whose estimates are constant across LLMs; this includes the zero-answer setting, where Pearson correlation is undefined. Using a reference estimated with the prior, or restricting the measure to the $50$ skills with at least $20$ pool items, changes the agreement values but preserves the ordering of the compared methods.

\textbf{Per-LLM AUC.} Test-item AUC changes little in this setting, from $0.68$ with no answers to $0.69$ with all answers. The zero-answer prediction uses the same profile for every LLM, so reaching this AUC does not require estimating differences between their skills. We therefore use benchmark-score error and profile agreement to assess what additional answers reveal about each LLM.

\begin{table}[t]
\centering
\small
\caption{Profiling $200$ held-out LLMs from $N$ answers. Error: mean absolute benchmark-score error over LLMs and benchmarks. Agreement: mean per-skill Pearson correlation with profiles estimated from all pool answers without the prior. AUC: mean per-LLM test-item AUC. Both selection methods are evaluated up to $500$ answers; random-selection results are averaged over three seeds. The all-answers error and AUC use the estimator with the prior.}
\label{tab:profiling}
\setlength\tabcolsep{4pt}
\begin{tabular}{rccccccc}
\toprule
 & \multicolumn{3}{c}{\textbf{Error}} & \multicolumn{2}{c}{\textbf{Agreement}} & \multicolumn{2}{c}{\textbf{AUC}} \\
\cmidrule(lr){2-4}\cmidrule(lr){5-6}\cmidrule(lr){7-8}
$N$ & Random & Adaptive & Count & Random & Adaptive & Random & Adaptive \\
\midrule
0 & 0.095 & 0.095 & 0.095 & 0.000 & 0.000 & 0.682 & 0.682 \\
5 & 0.087 & 0.079 & 0.202 & 0.207 & 0.301 & 0.682 & 0.684 \\
10 & 0.082 & 0.068 & 0.218 & 0.252 & 0.351 & 0.682 & 0.686 \\
20 & 0.075 & 0.061 & 0.217 & 0.329 & 0.417 & 0.682 & 0.687 \\
50 & 0.068 & 0.056 & 0.154 & 0.396 & 0.470 & 0.684 & 0.688 \\
100 & 0.061 & 0.053 & 0.107 & 0.451 & 0.532 & 0.686 & 0.689 \\
200 & 0.055 & 0.048 & 0.075 & 0.502 & 0.595 & 0.688 & 0.690 \\
500 & 0.049 & 0.046 & 0.052 & 0.570 & 0.655 & 0.690 & 0.691 \\
\midrule
All $7{,}618$ & 0.042 & & & & & 0.693 & \\
\bottomrule
\end{tabular}
\end{table}

\section{Per-Benchmark Routing}
\label{app:perbench}

Table~\ref{tab:perbench} expands the per-benchmark routing breakdown summarised in \S\ref{sec:exp:routing}. SkillEval beats the strongest-model baseline and IRT 2PL on every benchmark, with the largest absolute gains over IRT on MuSR ($+12.5$ pts) and IFEval ($+14.0$ pts).

\begin{table}[t]
\centering
\small
\caption{Routing Acc@1 by source benchmark on the held-out test items.
\textbf{Strongest} is the train-selected top-1 LLM. SkillEval beats the
single-LLM baseline and IRT 2PL on all five benchmarks.}
\label{tab:perbench}
\begin{tabular}{lrcccc}
\toprule
\textbf{Benchmark} & $n$ & \textbf{Random} & \textbf{Strongest} & \textbf{IRT 2PL} & \textbf{SkillEval} \\
\midrule
MATH    & 264  & 0.152 & 0.375 & 0.530 & \textbf{0.542} \\
BBH     & 1159 & 0.471 & 0.716 & 0.690 & \textbf{0.735} \\
GPQA    & 246  & 0.280 & 0.370 & 0.402 & \textbf{0.407} \\
MuSR    & 136  & 0.301 & 0.529 & 0.419 & \textbf{0.544} \\
IFEval  & 100  & 0.450 & 0.810 & 0.680 & \textbf{0.820} \\
\midrule
Overall & 1905 & 0.405 & 0.616 & 0.611 & \textbf{0.657} \\
\bottomrule
\end{tabular}
\end{table}

\section{Weak-Beats-Strong Inversions}
\label{app:wbs}

\begin{figure}[t]
\centering
\includegraphics[width=0.85\linewidth]{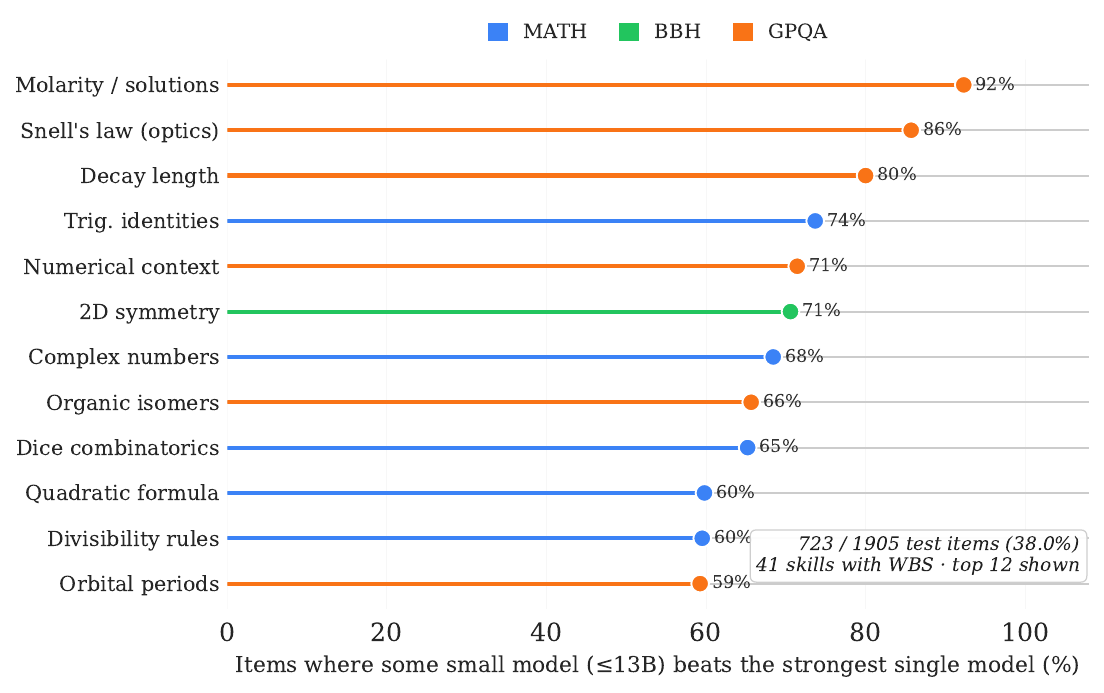}
\caption{Distribution of ``weak-beats-strong'' items by source benchmark.
$38\%$ of held-out items ($n{=}723/1{,}905$) admit at least one
$\leq{}13$B LLM that out-performs the global strongest model, spanning
$41$ skill clusters: BBH $329$, MATH $156$, GPQA $155$, MuSR $64$,
IFEval $19$.}
\label{fig:wbs}
\end{figure}

Figure~\ref{fig:wbs} expands the per-benchmark distribution of
``weak-beats-strong'' inversions summarised in \S\ref{sec:exp:routing}.
Across the $1{,}905$ held-out test items, $723$ ($38\%$) admit at least
one LLM with $\leq{}13$B parameters that out-performs the global
strongest model, spanning $41$ distinct skill clusters. The phenomenon
is broad rather than driven by a single benchmark: BBH contributes
$329$, MATH $156$, GPQA $155$, MuSR $64$, IFEval $19$. This is the
headroom that skill-conditional routing exploits and scalar-ability
routers cannot reach.

\section{Cross-Benchmark Transfer}
\label{app:transfer}

To probe whether the learned skill space generalises across benchmark families,
we hold out two entire benchmarks (MuSR and IFEval) and train SkillEval on
the remaining three (MATH, BBH, GPQA), using the same $K{=}100$ HAC clusters,
$M{=}3{,}811$ LLMs, and $15$-epoch Adam schedule as the main run. The
training set covers $8{,}231$ items; the held-out evaluation set covers
$1{,}292$ items.

Aggregate diagnostic fit on the held-out items collapses to AUC $0.469$
(test accuracy $0.460$), indicating that on combined unseen-benchmark items
the model is no better than chance. The picture sharpens when split by
benchmark (Table~\ref{tab:transfer}). On MuSR, where the multi-step reasoning
skills overlap with BBH, transfer is intact: SkillEval reaches Acc@1 $0.470$,
matching the strongest-model baseline ($0.467$) but below the
full-training reference ($0.544$). On IFEval, however, SkillEval drops to
$0.410$, below random ($0.437$) and $32$ points below the strongest model
($0.731$).

\paragraph{Skill coverage in the transfer experiment.}
With MuSR and IFEval held out, $15$ of the $100$ skills have no training item. These skills were defined during Q-matrix construction on the full item bank, but their mastery parameters receive no response-based training in this experiment. IFEval items are assigned to $62$ skills, $14$ of which are among these $15$, and $23\%$ of IFEval items involve at least one of them; for MuSR the share is below $1\%$. This coverage gap is consistent with the larger failure on IFEval, but it does not explain it fully, since most IFEval items involve only skills that were trained. We therefore report the IFEval result as a limit of cross-benchmark transfer: deployments should include the target benchmark family in the training mixture rather than rely on zero-shot transfer.

\begin{table}[t]
\centering
\small
\caption{Cross-benchmark transfer with MuSR and IFEval held out entirely.
``Reference'' is the full-training Acc@1 from \S\ref{sec:exp:routing}.
Transfer succeeds on MuSR (overlap with BBH multi-step reasoning) but fails on
IFEval, where $14$ of the $62$ assigned skills receive no training items.}
\label{tab:transfer}
\begin{tabular}{lrcccc}
\toprule
\textbf{Held-out} & $n$ & \textbf{Random} & \textbf{Strongest} & \textbf{SkillEval (transfer)} & \textbf{Reference} \\
\midrule
MuSR    & 756  & 0.415 & 0.467 & 0.470 & 0.544 \\
IFEval  & 536  & 0.437 & 0.731 & 0.410 & 0.820 \\
\midrule
Combined & 1292 & 0.400 & 0.577 & 0.445 & --- \\
\bottomrule
\end{tabular}
\end{table}

\section{Controls for the Validity of $\Theta$}
\label{app:validity_controls}

Section~\ref{sec:eval_theta} reports the correlation between $\Theta$ and held-out per-skill accuracy. Such a correlation could arise if $\Theta$ only reflected how strong an LLM is overall. For each skill $k$ we therefore compute the partial correlation, across LLMs, between the mastery $\theta_k$ and the held-out accuracy on skill $k$ after regressing out the LLM's overall training accuracy. Over the $41$ skills with at least ten held-out items, the mean partial correlation is $0.35$, averaged over five seeds. For a descriptive null comparison, we fix the seed-$42$ mastery estimates and permute held-out item-to-skill assignments within each benchmark, preserving each skill's held-out size and benchmark composition; the mean partial correlation over $20$ such permutations is $0.22$. Controlling instead for the LLM's five per-benchmark training accuracies gives $0.135$, against a placebo mean of $-0.02$. Controlling for the finer profile of per-subtask training accuracies ($38$ subtasks) reduces the mean partial correlation to $0.021$, indicating substantial overlap with subtask structure. These comparisons indicate that $\Theta$ carries information beyond overall and benchmark-level accuracy; they do not separate it from subtask-level accuracy.

\section{Empirical Co-Mastery Patterns}
\label{app:comastery_details}

\paragraph{Mastery and link rule.}
This exploratory analysis asks whether LLMs that master one skill also tend to master another, even when the reverse is less common. We find such asymmetric co-mastery patterns across most skills, with similar patterns visible in the observed responses. In Equation~\ref{eq:ncdm}, estimated mastery $\theta$ is interpreted relative to item difficulty, so its value alone does not specify expected accuracy on a skill's items. For this analysis, we classify an LLM as mastering a skill when its predicted probability of a correct answer, averaged over the items assigned to that skill, exceeds $0.5$, that is, when the model expects it to answer more than half of those items correctly.

For two skills $A$ and $B$, let $P(A\mid B)$ denote the share of LLMs mastering $A$ among those mastering $B$. For example, $52$ of the $55$ LLMs mastering \emph{combinatorics of dice rolls} also master \emph{divisibility rules for factorials}, so $P(\text{divisibility}\mid\text{dice})=52/55\approx0.95$. In the reverse direction, only $52$ of the $167$ LLMs mastering divisibility rules also master dice-roll combinatorics, so $P(\text{dice}\mid\text{divisibility})=52/167\approx0.31$. Almost every LLM that masters the dice-roll skill thus also masters the divisibility skill, whereas most LLMs that master the divisibility skill do not master the dice-roll skill. We represent this pattern by a link from divisibility rules to dice rolls.

We draw a link from $A$ to $B$ when two conditions hold. First, we require \mbox{$P(A\mid B)$}${}-{}$\mbox{$P(B\mid A)$}${}>0.3$, capturing an asymmetry like the one above. Second, we require a strong association. The odds of an event are its probability divided by the probability that it does not occur, so a probability of $0.75$ corresponds to odds of $3$. The odds ratio compares the odds of mastering $A$ among LLMs that master $B$ with those among LLMs that do not. A ratio of $1$ indicates no association; a ratio above $1$ means mastery of $A$ is more likely among LLMs mastering $B$. We require an estimated odds ratio above $10$. Without this condition, unrelated skills could be linked simply because one is mastered more frequently than the other.

\paragraph{Odds ratio.}
For two skills $A$ and $B$, let $n_{AB}$ be the number of LLMs that master both, $n_{A\bar{B}}$ the number that master $A$ but not $B$, $n_{\bar{A}B}$ the number that master $B$ but not $A$, and $n_{\bar{A}\bar{B}}$ the number that master neither. The Haldane correction adds $0.5$ to each count, which keeps the odds ratio finite when a count is zero:
\begin{equation*}
\mathrm{OR}(A,B) \;=\; \frac{(n_{AB}+0.5)\,(n_{\bar{A}\bar{B}}+0.5)}{(n_{A\bar{B}}+0.5)\,(n_{\bar{A}B}+0.5)}.
\end{equation*}

\paragraph{Co-mastery graph.}
Applying this rule to every pair of skills gives a \emph{co-mastery graph} with skills as nodes. The graph orders two skills when following links in their direction leads from one to the other. To simplify the graph, we remove a direct link when the original graph also connects the same pair through one intermediate skill. This leaves $304$ links connecting $86$ of the $100$ skills; some retained links may still be implied by longer paths. Of these $86$ skills, $18$ have no incoming link and $13$ have no outgoing link. Every link points from a more commonly mastered skill to a less commonly mastered one, because $P(A\mid B)$ can exceed $P(B\mid A)$ only when more LLMs master $A$ than $B$. Its direction therefore follows from the construction and does not show that $A$ must be learned before $B$. Repeating the construction and link removal with a mastery cut-off of $0.4$ instead of $0.5$ gives $391$ links connecting $89$ skills, and with $0.6$ it gives $209$ links connecting $68$ skills. Only $29$ direct links appear at all three cut-offs, so individual links are sensitive to the cut-off.

\paragraph{Checks.}
Two checks support these patterns. First, we construct a second graph using the same LLMs, items and skill assignments, but classify an LLM as mastering a skill if it answers more than half of that skill's items correctly. This graph orders $1{,}325$ pairs. The co-mastery graph orders $983$ in the same direction and one in the opposite direction, leaving the rest unordered. Wherever both graphs order a pair, they agree in all but one case. Second, we independently shuffle each skill's mastery labels across LLMs, preserving the share mastering each skill while disrupting the observed associations. Before link removal, the rule finds at most $51$ links in any of $200$ shuffles, compared with $1{,}031$ in the original profiles. This supports associations beyond individual skill frequencies. Across the $304$ retained links, the median of $P(A\mid B)$ is $0.87$ and the median of $P(B\mid A)$ is $0.26$. A benchmark average does not expose these relationships between named skills. For dataset design, the graph can suggest skills to assess within groups of LLMs sharing a strength, as dice-roll combinatorics distinguishes among the $167$ LLMs that master divisibility rules.

\paragraph{Cross-family overlap.}
We also apply the link rule separately to each model family with at least $100$ LLMs: Llama, Qwen, Mistral, Gemma and Phi. We compare the resulting link sets before removing any links. For two families, the Jaccard index is the number of links shared by both divided by the number found in at least one. Across the ten family pairs, it ranges from $0.09$ to $0.31$. The detected links are therefore only partly shared across the sampled families; whether these patterns persist in other LLM populations remains to be tested.

\section{Benchmark Redundancy Details}
\label{app:redundancy}

\begin{figure}[t]
\centering
\includegraphics[width=0.6\linewidth]{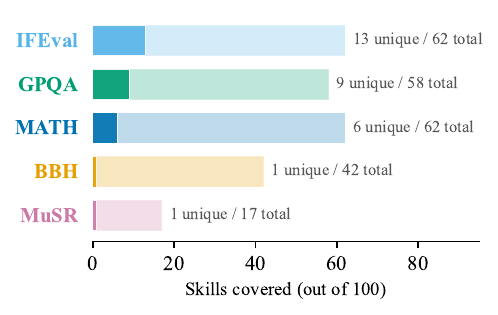}
\caption{Skill coverage per benchmark. Dark segments show skills unique
to that benchmark; light segments show shared skills. IFEval contributes
$13$ unique skills, while BBH has only $1$. MATH~+~IFEval together cover
$87/100$ skills.}
\label{fig:upset}
\end{figure}

Figure~\ref{fig:upset} shows skill coverage per benchmark: dark segments
mark skills unique to that benchmark, light segments mark skills shared
with at least one other benchmark. IFEval contributes the most unique
skills ($13$ of $62$), while BBH contributes only $1$ unique skill of
$42$. MATH and IFEval together cover $87$ of $100$ curated skills.

\begin{figure}[t]
\centering
\includegraphics[width=0.55\linewidth]{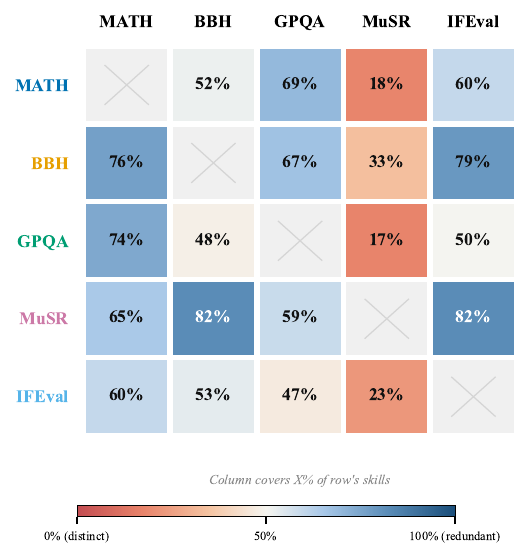}
\caption{Pairwise skill coverage matrix. Cell $(i,j)$: fraction of row
benchmark's skills also covered by column benchmark. Red = low overlap
(distinct), blue = high overlap (redundant). MuSR's skills are $82\%$
covered by BBH and IFEval.}
\label{fig:redmat}
\end{figure}

Figure~\ref{fig:redmat} shows the pairwise coverage matrix: cell
$(i,j)$ reports the fraction of benchmark~$i$'s skills also covered by
benchmark~$j$. MuSR is the most redundant row: $82\%$ of its skills are
covered by BBH, and $82\%$ by IFEval. IFEval is the most independent row, with the
lowest average overlap with the other benchmarks.

\section{Differences Between Base and Instruction-Tuned Models: Details}
\label{app:alignment}

\begin{figure}[t]
\centering
\includegraphics[width=\linewidth]{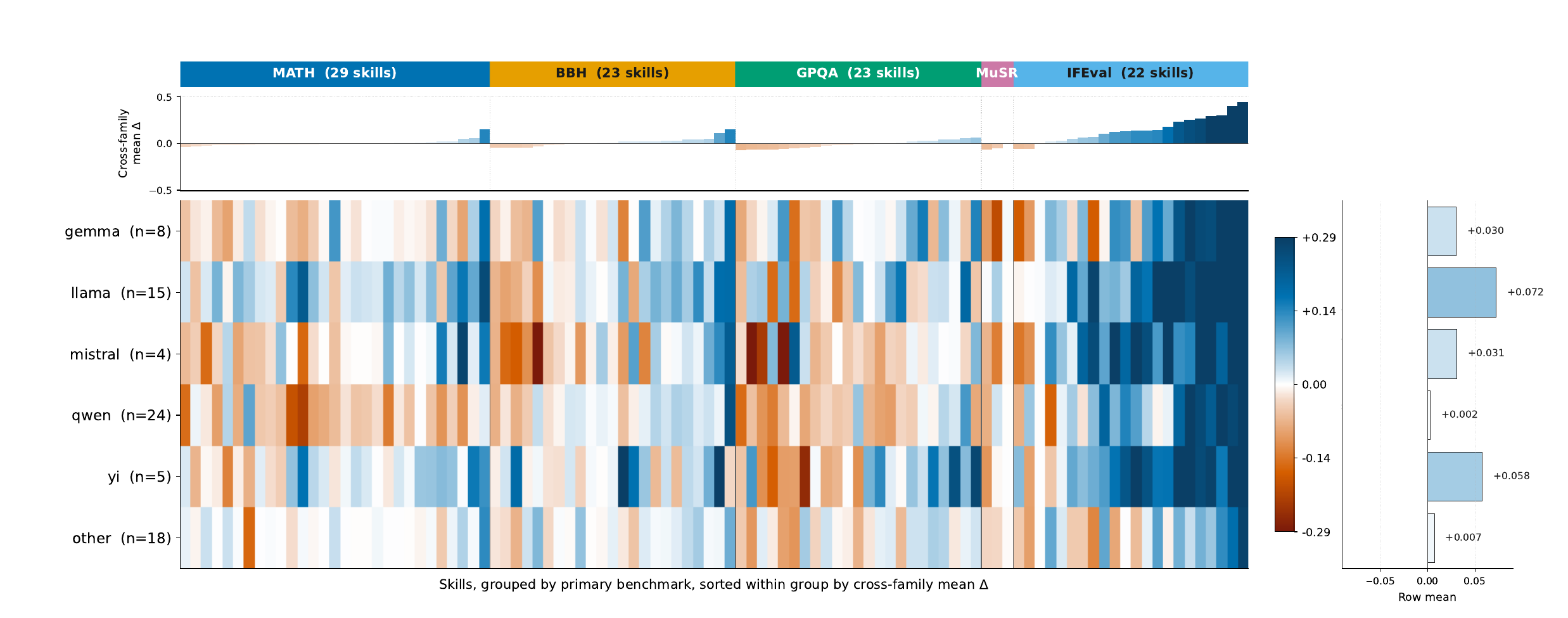}
\caption{Mean estimated mastery difference (instruction-tuned$-$base) per model family and benchmark.
Blue = improved, red = degraded. The IFEval column is positive for every evaluated family;
the MuSR column is negative for every evaluated family.}

\label{fig:alignfam}
\end{figure}

Figure~\ref{fig:alignfam} reports estimated mastery differences by model family and benchmark, with each skill assigned to the benchmark that contributes most of its items. Mean differences over IFEval skills are positive for every evaluated family.
The MuSR column is uniformly negative, and the MATH/GPQA columns
are near-zero or slightly negative for most families. Llama shows the broadest
gains; Qwen is essentially flat.

\section{Limitations}
\label{app:limitations}

\paragraph{Scope of the new-LLM experiments.}
The new-LLM results use $200$ of the $763$ held-out LLMs and one trained network. We evaluate one adaptive selection rule at answer budgets up to $500$. The prior is estimated from profiles produced by that network, while observed scores for the smaller benchmarks are based on relatively few test items. The results describe the evaluated LLMs, network and item pool; they do not establish an advantage for adaptive selection in every setting.

\paragraph{Co-mastery vs.\ causal prerequisites.}
The co-mastery graph in Appendix~\ref{app:comastery_details} is constructed from
statistical co-mastery ($\mathrm{OR}>10$), not causal dependency. The
observed structure likely reflects a mix of skill-difficulty ordering
and shared training-data signal rather than experimentally verifiable
cognitive prerequisites. Establishing causal prerequisite relations
would require intervention experiments (skill-specific fine-tuning)
beyond our scope; we therefore phrase all structural claims in terms
of co-mastery.

\paragraph{Q-matrix inter-annotator agreement.}
The Q-matrix is constructed semi-automatically via LLM-proposed skill
phrases clustered by SBERT and HAC. Cluster choice is validated by
downstream NCDM AUC rather than by human adjudication. A human validation of a sample of skill--question assignments is reported in Section~\ref{sec:eval_skill_quality}. It tests whether an assigned skill is needed, not whether needed skills are missing. Items near cluster boundaries should be interpreted with
additional caution. We sweep the number of skill
clusters $K$ and report sensitivity to $K$, but we do not separately
vary the clustering algorithm (HAC), the skill-extraction prompt, or
the sentence-embedding source (SBERT); we leave a joint pipeline-robustness
study to future work.

\paragraph{Evaluation scope.}
Our item-wise OOD setting holds out all responses to the test items while retaining the same benchmarks and LLM population. We treat the Q-matrix as part of the test design: the skill taxonomy and Q-matrix are constructed before this split, using question text and benchmark/subtask metadata from the complete item bank, including test items. Skill extraction and clustering use item content rather than the evaluated LLMs' responses. However, the clustering method and the number of clusters were chosen with reference to downstream performance (Appendices~\ref{app:qmatrix} and~\ref{app:ksweep}). Item difficulty and discrimination for test items are predicted from their text by projections trained on training items only. The evaluation is therefore transductive with respect to the skill representation, while test-item responses are excluded from parameter fitting: it measures prediction of held-out responses within a fixed, item-bank-derived taxonomy. Generalization remains limited. IRTNet outperforms \textsc{SkillEval} on held-out-item prediction (Table~\ref{tab:item_prediction}). When MuSR and IFEval are jointly excluded from response-based training, observed routing accuracy on MuSR is similar to that of the single-model baseline selected by training accuracy ($0.470$ versus $0.467$), whereas IFEval routing falls below the random-routing baseline ($0.410$ versus $0.437$; Appendix~\ref{app:transfer}). These results limit the scope of our routing and profiling claims; reliable transfer to new benchmark domains or skill families remains to be established.

\paragraph{Benchmark coverage.}
SkillEval covers five English-language benchmarks (MATH, BBH, GPQA, MuSR, IFEval). Notable gaps include multilingual reasoning, tool use,
long-context retrieval, safety/red-team benchmarks, and open-ended
generation. The Q-matrix pipeline is designed to extend to new
benchmarks without retraining existing skill clusters: adding a new
benchmark requires running the LLM skill extraction pipeline on its
items, assigning each extracted skill to its nearest existing cluster
(or expanding $K$), and retraining NCDM.
\end{document}